\documentclass[review]{elsarticle} %review  or  5p

\usepackage{hyperref}
\usepackage{changepage}
\journal{International Journal of Hydrogen Energy}
\usepackage{lineno}
\usepackage{amsmath} %added for maths environments (equation and align)
\usepackage{amssymb}
\usepackage{upgreek}
\usepackage{bm} %bold symbols by using \bm
\usepackage{mathtools, nccmath} %added for \xrightleftharpoons
\biboptions{sort&compress}
\usepackage{makecell}  
\usepackage{float} % To use H in floating figures
\usepackage{tablefootnote}
\usepackage{subcaption} %allowing for subcaptions and subfigures
\usepackage{algorithm} %added for algorithm box
\usepackage{algpseudocode}
\usepackage[margin=2.5cm]{geometry}% TO PLAY WITH THE MARGINS
\usepackage[capitalise, nameinlink]{cleveref}

\crefname{appendix}{}{}

\usepackage{placeins}

\usepackage{booktabs}
\usepackage{multicol}
\usepackage{multirow}

\usepackage{todonotes} %added for todo notes
\let\oldtodo\todo
\renewcommand{\todo}[1]{\oldtodo[inline]{#1}}

\usepackage[normalem]{ulem}

\begin{document}

\begin{frontmatter}
\title{A neural network machine-learning approach for characterising hydrogen trapping parameters from TDS experiments}

\author[1]{Nicoletta Marrani}

\author[1]{Tim Hageman\corref{A1}}
\cortext[A1]{Corresponding authors}
\ead{tim.hageman@eng.ox.ac.uk}

\author[1]{Emilio Martínez-Pañeda\corref{A1}}
\ead{emilio.martinez-paneda@eng.ox.ac.uk}

\address[1]{Department of Engineering Science, University of Oxford, Oxford OX1 3PJ, UK}

\begin{abstract} %max 150 words
\noindent The hydrogen trapping behaviour of metallic alloys is generally characterised using Thermal Desorption Spectroscopy (TDS). However, as an indirect method, extracting key parameters (trap binding energies and densities) remains a significant challenge. To address these limitations, this work introduces a machine learning-based scheme for parameter identification from TDS spectra. A multi-Neural Network (NN) model is developed and trained exclusively on synthetic data to predict trapping parameters directly from experimental data. The model comprises two multi-layer, fully connected, feed-forward NNs trained with backpropagation. The first network (classification model) predicts the number of distinct trap types. The second network (regression model) then predicts the corresponding trap densities and binding energies. The NN architectures, hyperparameters, and data pre-processing were optimised to minimise the amount of training data. The proposed model demonstrated strong predictive capabilities when applied to three tempered martensitic steels of different compositions. The code developed is freely provided. \\
\end{abstract}

% 3-5 Highlights (max 85 char per highlight), optional
% \begin{highlights}
% \item A
% \item B
% \item C
% \item D
% \item E
% \end{highlights}


\end{frontmatter}

\section{Introduction}
\label{sec:intro}
%\linenumbers

To transition away from fossil fuels as the primary energy source, the global energy industry has increasingly focused on developing low-carbon technologies. Alongside the widespread adoption of renewable energy, hydrogen is gaining prominence as both a fuel and an energy carrier  \citep{moriarty_prospects_2019, Genovese2023, Liu2021}. Its natural abundance and low projected environmental impact position hydrogen as a promising solution for decarbonising traditionally hard-to-abate industries \citep{pleshivtseva_comprehensive_2023, guo_hydrogen_2024}. However, the development of a hydrogen-based economy is hindered by hydrogen’s tendency to degrade the mechanical properties of structural materials \textemdash a phenomenon known as hydrogen embrittlement \citep{chen_hydrogen_2025}. A detailed understanding of hydrogen embrittlement is necessary before existing energy infrastructure, such as gas pipelines, is adapted to transport hydrogen \citep{Mandal2024, Wijnen2025}, with one of the main limitations being the characterisation of metal properties.

Despite extensive experimental and computational efforts to understand the mechanisms of hydrogen embrittlement, characterising the properties of metals and predicting their degradation due to hydrogen remains a challenge \citep{paneda_progress_2021}. These predictions require a thorough examination of hydrogen-metal interactions, including ingress \citep{Hageman2022a, Hageman2023b, Schwarzer2022, Cupertino-Malheiros2024}, diffusion through the lattice \citep{Ferrin2012, zafra_relative_2023}, and trapping at microstructural imperfections \citep{ Lu2022, cupertino2023hydrogen,santos_maldonado_influence_2024}. In particular, the rapid diffusion of hydrogen through the metal lattice and its subsequent trapping at crystal defects have been identified as key factors influencing susceptibility to hydrogen embrittlement \citep{chen_hydrogen_2025, turnbull_perspectives_2015}.

One well-established technique for characterising hydrogen trapping behaviour in metallic alloys is Thermal Desorption Spectroscopy (TDS) \citep{verbeken_analysing_2012}, also referred to as Thermal Desorption Analysis (TDA). TDS analysis has been used extensively to understand how microstructural features\textemdash such as vacancies \citep{nagumo_predominant_2019}, dislocations \citep{depover_thermal_2018, l_hydrogen_2025, shang_effects_2025}, grain boundaries \citep{choo_thermal_1982}, voids \citep{van_veen_hydrogen_1988, lee_hydrogentrappingby_nodate, yaktiti_hydrogen_2022}, and precipitates \citep{depover_effect_2016, wei_quantitative_2006, wei_hydrogen_2003}\textemdash interact with and retain hydrogen. Each of these features, commonly referred to as traps, is characterised by a unique binding energy and trap density, which TDS can quantify. The technique is valuable for comparing hydrogen trapping across materials and examining the influence of environmental and processing factors, such as variations in hydrogen charging conditions \citep{silverstein_hydrogen_2018} and tempering temperature \citep{kim_effect_2023, depover_thermal_2018}. Additionally, TDS can be combined with other techniques, such as X-ray diffraction and microstructural analysis, to investigate hydrogen trapping mechanisms in different crystal structures \citep{silverstein_mechanisms_2017}.

Although TDS is a powerful tool for probing trapping properties, it is an indirect method, and extracting critical parameters, such as trap binding energies and densities, requires careful post-processing and interpretation of the desorption spectrum \citep{hurley_numerical_2015}. As TDS obtains the desorption spectrum of all the traps combined, this combined spectrum needs to be split into contributions of individual traps \citep{verbeken_analysing_2012}. Kissinger’s first-order rate theory \citep{kissinger_variation_1956} is the simplest framework for evaluating TDS spectra, commonly applied through the Choo-Lee approach \citep{choo_thermal_1982}. While Kissinger's method is popular for its simplicity, it has limitations, primarily due to assumptions of infinitely fast hydrogen diffusion and neglecting the role of multiple interacting trapping mechanisms. Consequently, it is only valid for TDS tests on fast diffusion materials (e.g., bcc alloys), using very thin samples, and at very slow heating rates \citep{raina_analysis_2018}.

An alternative framework, proposed by McNabb and Foster \citep{mcnabb_new_1963}, accounts for the role of trapping and de-trapping in hydrogen diffusion. It assumes sparsely distributed, isolated traps and requires numerical methods for solving the system of coupled differential equations. This framework can be simplified using Oriani’s approach \citep{oriani_diffusion_1970}, which assumes equilibrium between hydrogen in the lattice and trapping sites, and which generally applies, especially for low trap occupancies \citep{bombac_theoretical_2017}. A number of studies have used these hydrogen diffusion-trapping models, using finite difference  \citep{drexler_critical_2021} and finite element methods \citep{legrand_towards_2015, raina_analysis_2018} to solve the resulting PDEs, considering both single and multi-trap scenarios \citep{garcia-macias_tds_2024, song_theory_2013, turnbull_modelling_1997, kirchheim_bulk_2016}. Garcia-Macias et al. \citep{garcia-macias_tds_2024} developed \texttt{TDS Simulator}, a MATLAB App capable of modelling TDS experiments for any material and test parameters, and extracting trapping characteristics (the number of traps, binding energies, and densities) from experimental desorption curves. They implemented the McNabb-Foster and Oriani frameworks for multiple trap sites, using a deterministic parameter inference algorithm based on particle swarm optimisation for data fitting. Delaporte-Mathurin et al. \citep{delaporte-mathurin_parametric_2021} developed a finite element code for hydrogen transport simulations. Using the Nelder-Mead algorithm and parametric optimisation, they determined the trapping characteristics of materials like Aluminium, Tungsten, EURO-FER, and Beryllium. This method successfully extracted trapping parameters from experimental TDS spectra. Drexler et al. \citep{drexler_model-based_2019} used a trap-diffusion finite element model-based evaluation procedure to extract the trapping quantities of interest from the TDS spectra of two Fe-C-Ti alloys. The developed optimisation routine was able to successfully predict both the trapping parameters, i.e. binding energy and density, and the initial hydrogen concentration of the two alloys. However, all these approaches have limitations: they require defining \textit{a priori} the number of traps, which could lead to unsatisfactory fits even if the algorithm converges. Additionally, as these methods rely on conventional optimisation algorithms, they require the initial estimates of the trapping energy and density to be close to the correct values, as otherwise the algorithm can converge to local minima, preventing the correct material parameters from being predicted. Finally, they require large amounts of simulations to be performed every time the parameters are fitted, with this simulation effort not being able to be reused for future predictions. This work aims to overcome these limitations using ML approaches.

Growth in computing power has led to the development of Artificial Intelligence (AI), a powerful tool now used in numerous fields, including natural language processing, image recognition, finance, medicine, and material science \cite{Carleo2019, sharifani_machine_2023, Karniadakis2021,garcia-merino_multielement_2023}. Machine learning (ML), a branch of AI, uses methods such as decision trees, linear regression, and artificial neural networks (ANNs) to extract knowledge from data, learn from it, and solve complex problems \cite{sharifani_machine_2023}. Recently, deep learning, a subset of ML, has emerged as a prominent tool, utilising deep ANNs to address highly complex issues. Deep learning models excel in analysing large datasets and are particularly effective when large amounts of data are available \cite{chauhan_review_2018}.

ANNs are increasingly used in material science and engineering \cite{bhadeshia_neural_1999, Bock2019, Jin2023}, and their use has recently been extended to the analysis of hydrogen-material interactions \cite{thankachan_artificial_2017,malitckii_evaluation_2020,malitckii_study_2020,fangnon_prediction_2023}. However, they are yet to be employed to infer trapping characteristics from TDS output. Due to the vast amount of training data that can easily be generated through numerical simulations, employing deep artificial neural networks is suitable for addressing the complexity of the interpretation of TDS spectra. The objective of this work is to introduce a machine learning-based scheme for parameter identification from TDS experiments. The aim is to have this ANN trained solely on simulated data, while being accurate in predicting parameters from experimental data. Hence, we present an ML-based framework to predict the number of trap types and the corresponding binding energies and densities from experimental TDS spectra, without needing to provide the number of different trap types and an initial estimate of the binding energies and densities. Furthermore, by using neural networks, training data can be generated a priori and reused across experiments to efficiently and quickly identify traps without needing any manual post-processing steps, accelerating the capabilities of TDS to characterise the hydrogen trapping properties of metals. 

The remainder of this paper is organised as follows. First, the hydrogen transport model is introduced in \cref{sec:model}, including the governing equations and its finite element method (FEM) implementation. Next, the machine learning methodology is described in \cref{section:ML_approach}, outlining the architecture of the multi-NN model, the model optimisation process, and the generation of training data using FEM simulations. Finally, the performance of the proposed multi-NN model approach is demonstrated through three case studies involving tempered martensitic steels of different composition in \cref{sec:results}. The paper concludes with a discussion of the approach limitations and main advantages.

\section{Hydrogen transport model}
\label{sec:model}

In order to generate the training data and verify the correctness of the ANN predictions, a numerical model based on the finite element method was developed to simulate TDS spectra for an arbitrary choice of material and test parameters. Upon ingress into the metal, hydrogen diffuses through the crystal lattice by hopping between normal interstitial lattice sites (NILS). As the hydrogen diffuses through the crystal lattice, it can be captured by micro-structural heterogeneities, known as traps. Thus, the total hydrogen concentration $C$ is the sum of the lattice hydrogen concentration $C_\text{L}$ and the trapped hydrogen concentration $C_\text{T}$. Within the finite element model, the lattice and trapped concentrations are used as degrees of freedom to be solved for. 

\begin{figure*}[t]
    \centering
    \begin{subfigure}{0.32\textwidth}
        \centering
        \includegraphics{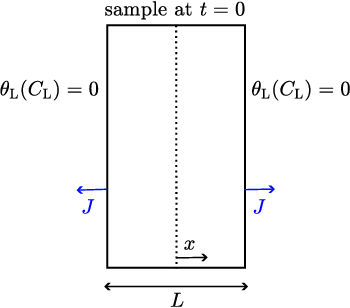}
        \caption{}
        \label{fig:schematic_geometry_model}
    \end{subfigure}
    \hfill
    \begin{subfigure}{0.32\textwidth}
        \centering
        \includegraphics{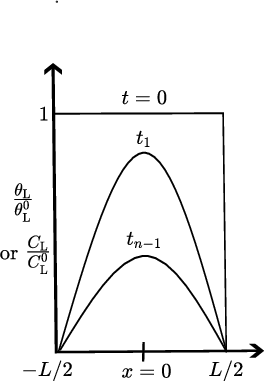}
        \caption{}
        \label{fig:schematic_flux_evolution}
    \end{subfigure}
    \hfill
    \begin{subfigure}{0.32\textwidth}
        \centering
        \includegraphics{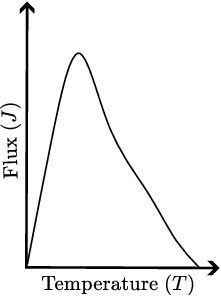}
        \caption{}
        \label{fig:schematic_spectrum}
    \end{subfigure}
    \caption{Thermal Desorption Spectroscopy (TDS) experiments. (a) Schematic illustration of the sample geometry, initial and boundary conditions in a TDS test. (b) Transient solution curves of the normalised lattice occupancy fraction $\theta_\text{L}/\theta_\text{L}^0$ (or $C_\text{L}/C_\text{L}^0$) along the sample thickness $L$ at different times $t$. (c) Schematic of a typical hydrogen desorption flux curve (flux versus temperature) obtained in a TDS test.}
    \label{fig:schematics_model}
\end{figure*}

In TDS experiments, the hydrogen desorption rate is measured from a plate of thickness $L$ that has been uniformly pre-charged to achieve a homogeneously distributed hydrogen concentration of $C_\text{L}^{0}$. After an initial rest period $t_\text{rest}$ (in which the sample is held at room temperature and transferred from the hydrogen charging device to the TDS setup), heating takes place at a constant heating rate $\phi$, causing the temperature of the sample to rise from $T_\text{min}$ to $T_\text{max}$ and desorption to occur as the trapped hydrogen in released. The temperature $T$ at any time $t$ during the TDS test is defined as,
\begin{equation}
    T=T_\text{min} + \phi \langle t - t_\text{rest} \rangle,
\end{equation}
where the Macaulay brackets $\langle \cdot \rangle$ ($\langle x \rangle = x$ if $x\geq0$, otherwise $\langle x \rangle = 0$) are used to indicate an initial period $t_\text{rest}$ where the temperature is held constant at $T_{min}$, the initial temperature ($293.15\;\text{K}$, typically). We assume that the heating rate $\phi$ is sufficiently slow compared to the rate of heat transfer, such that the temperature distribution $T(t)$ is assumed to be spatially uniform throughout the TDS sample. 
Given that the sample thickness is significantly smaller than all other dimensions, desorption primarily occurs at the surfaces corresponding to $x=\pm L/2$; consequently, the problem is effectively one-dimensional. As the temperature rises, the lattice hydrogen concentration $C_\text{L}$ evolves spatially and temporally as shown in \cref{fig:schematic_flux_evolution}. A flux profile similar to that in \cref{fig:schematic_spectrum} is obtained.

\begin{figure}
    \centering
    \includegraphics{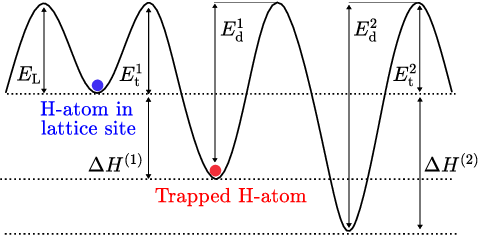}
    \caption{Overview of different energy levels associated with hydrogen diffusion in metals. $E_\text{L}$ is the interstitial lattice activation energy, $E_\text{t}\;(>0)$ is the trapping activation energy and $E_\text{d}\;(>0)$ is the de-trapping activation energy. It is generally assumed that $E_\text{t} \approx E_\text{L}$, and $E_\text{d}$ and $E_\text{t}$ define the binding energy $\Delta H\;(<0)$ \textit{via} $\Delta H = E_t - E_d$.}
    \label{fig:Definitions_Energies}
\end{figure}

\subsection{Governing Equations}

The diffusive transport of the hydrogen lattice concentration $C_\text{L}(x,t)$ over the NILS is dictated by Fick's Law, adjusted to account for the possibility of trapping and de-trapping. Assuming there are $\text{n}_\text{t}$ types of traps, Fick's second law can be written as: 
\begin{equation} \label{equation:diffusion}
       \frac{\partial C_\text{L}}{\partial t} + \sum_\text{i=1}^{n_\text{t}} \frac{\partial C_\text{T}^{(i)}}{\partial t} = D_\text{L} \frac{\partial^{2} C_\text{L}}{\partial x^2}
\end{equation}
where $C_\text{L}$ and $C_\text{T}$ are the lattice hydrogen concentration and the trapped hydrogen concentration, respectively, and $D_\text{L}$ is the lattice diffusion coefficient. This lattice diffusion coefficient depends on the temperature $T$ of the sample \textit{via} the following Arrhenius law:
\begin{equation}
    D_\text{L} = D_\text{0} \exp\left(-\frac{E_\text{L}}{RT}\right)
\end{equation}
with $D_0$ is the pre-exponential factor for lattice diffusion, 
$E_\text{L}$ is the activation energy of the lattice sites (in $\text{J}/\text{mol}$), and $R$ the universal gas constant. Eq. (\ref{equation:diffusion}) can be reformulated considering the standard definitions of density of lattice sites, $N_L$, density of trapping sites for a trap type, $N_T^{(i)}$, lattice occupancy, $0\leq\theta_\text{L}\leq1$, and trap occupancy, $0\leq\theta_\text{T}\leq1$. Accordingly, the lattice and trap hydrogen concentrations can be defined as $C_\text{L}=\theta_\text{L}N_\text{L}$ and $C_\text{T}^{\text{i}}=\theta_\text{T}^{i}N_\text{T}^{i}$, and the hydrogen transport equation can be written as,
\begin{equation} \label{equation:diffusion_fractions}
       \frac{\partial\theta_\text{L}}{\partial t} + \sum_{i=1}^{n_\text{t}} \frac{N_\text{T}^{(i)}}{N_\text{L}}\frac{\partial\theta_\text{T}^{(i)}}{\partial t} = D_\text{L} \frac{\partial^{2}\theta_\text{L}}{\partial x^2}
\end{equation}
The rate of trapped hydrogen, the $\partial\theta_\text{T}/\partial t$ term in \cref{equation:diffusion_fractions}, can be defined using the trapping and de-trapping rates, using the McNabb-Foster model, or directly linked to the lattice concentration by assuming equilibrium through the Oriani model. Here, and in our machine-learning approach, we will consider both. 

\subsubsection{McNabb-Foster model}

The McNabb-Foster model \citep{mcnabb_new_1963} accounts for trapping kinetics and, in combination with Eq. (\ref{equation:diffusion_fractions}), for the role of hydrogen diffusion. The framework assumes the presence of sparsely distributed and isolated traps, thus negating the possibility of any trap interaction from occurring. For hydrogen to transition from one trap site to another, it must pass through an intermediate lattice site. Modelling of the trapping and de-trapping kinetics of the hydrogen atoms is based on the energy landscape for the diffusion of hydrogen in metals (\cref{fig:Definitions_Energies}). The net rate of trapped hydrogen concentration is described as:
\begin{equation} \label{equation:mcnabb_foster_trapping}
       \frac{\partial\theta_\text{T}^{(i)}}{\partial\text{t}} = k^{(i)}\theta_\text{L}(1-\theta_\text{T}^{(i)}) - p^{(i)}\theta_\text{T}(1-\theta_\text{L}^{(i)})
\end{equation}
where $k$ and $p$ are the rates of hydrogen migrating from a NILS to a trap site and vice versa. They are defined as:
\begin{equation} \label{equation:mcnabb_foster_k}
       k^{(i)}=\nu_\text{t}^{(i)}\exp{\left(-\frac{E_\text{t}^{(i)}}{RT} \right)}
\end{equation}
\begin{equation} \label{equation:mcnabb_foster_p}
       p^{(i)}=\nu_\text{d}^{(i)}\exp{\left(-\frac{E_\text{d}^{(i)}}{RT} \right)}
\end{equation}
where $\nu_\text{t}^{(i)}$ and $\nu_\text{d}^{(i)}$ are the pre-exponential factors for trapping and de-trapping, respectively, and $E_\text{t}^{(i)}$ and $E_\text{d}^{(i)}$ are the trapping and de-trapping energies. 

\subsubsection{Oriani's model}

The McNabb-Foster framework can be simplified by assuming local equilibrium between trap and lattice sites, as demonstrated by Oriani \citep{oriani_diffusion_1970}. This assumption allows us to use the trap kinetic differential equation, \cref{equation:mcnabb_foster_trapping}, by assuming equilibrium to directly relate the trapped and lattice hydrogen:
\begin{equation} \label{equation:oriani_equilibrium}
    \frac{\theta_\text{T}^{(i)}}{1-\theta_\text{T}^{(i)}} = \frac{\theta_\text{L}}{1-\theta_\text{L}}K_\text{T}^{(i)}
\end{equation}
where $K_\text{T}$ is the equilibrium constant, which is given by
\begin{equation} \label{equation:oriani_equilibrium_constant}
    K_\text{T}^{(i)} = \frac{k^{(i)}}{p^{(i)}} = \frac{v_\text{t}^{(i)}}{v_\text{d}^{(i)}} \exp\left(\frac{-\Delta H^{(i)}}{RT}\right)
\end{equation}
with the binding energy being equal to the difference in trapping and de-trapping energies, $\Delta H^{(i)} = E_\text{t}^{(i)}-E_\text{d}^{(i)}$. As a result, \cref{equation:diffusion_fractions} can be reduced to a single second-order PDE solely dependent on $\theta_\text{L}$:
\begin{align}
    \frac{\partial \theta_\text{L}}{\partial t} \left( 1 + \sum_i \frac{N_\text{T}^{(i)} K_\text{T}^{(i)}}{N_\text{L} \left[ 1 + (K_\text{T}^{(i)} - 1) \theta_\text{L} \right]^2} \right) \nonumber \\ 
   \nonumber \\
    + \sum_i \frac{N_\text{T}^{(i)} K_\text{T}^{(i)} \Delta H^{(i)} \phi(\theta_\text{L} - \theta_\text{L}^2)}{N_\text{L} \left[ 1 + (K_\text{T}^{(i)} - 1) \theta_\text{L} \right]^2} 
    = D_\text{L} \frac{\partial^2 \theta_\text{L}}{\partial x^2} \label{equation:oriani_pde}
\end{align}

Both the Oriani and McNabb-Foster models typically provide the same results \citep{oriani_physical_1993}. This is further investigated in \cref{subsubsection:verification_equilibirumvalidity}, where it is shown that both models deliver the same result for relevant values of the vibration frequency. Oriani's model does not require solving for the trap density explicitly, halving the number of degrees of freedom. On the other hand, McNabb-Foster's relations bring a reduced nonlinearity, making them easier to solve. Both approaches are considered here, although the experimental case studies will only make use of the McNabb-Foster formulation for simplicity, considering a vibration frequency equal to the Debye frequency ($\nu=10^{13}\;\text{Hz}$), which delivers results identical to those obtained with Oriani's model (see \cref{subsubsection:verification_equilibirumvalidity}).

\subsubsection{Initial and boundary conditions}
On the left and right edges of the TDS specimen, interstitial lattice hydrogen is able to desorb from the sample, as shown in \cref{fig:schematic_geometry_model}. At equilibrium, the occupancy of hydrogen at these boundaries can be determined based on the exterior hydrogen pressure through Sievert's Law:
\begin{equation}
    \theta_\text{L} = \frac{S}{N_\text{L}}\sqrt{p_{\text{H}_2}}
\end{equation}
with the solubility $S$ depending on the temperature through an Arrhenius law. Here, we will idealise this to assume a full vacuum around the TDS sample (any hydrogen that diffuses outside of the metal is removed instantly), such that our boundary conditions can be simplified to $\theta_\text{L}=0$. This is representative of modern ultra-high-vacuum TDS equipment. To enforce this and obtain the hydrogen flux, the boundary flux (per unit area) is defined through a penalty-based approach, 
\begin{equation} \label{equation:peanlty_bc}
    j  =\left. -D_\text{L} \frac{\partial C_\text{L}}{\partial x}\right|_{x=L/2}  = k\theta_\text{L}\exp\left(\frac{-E_\text{bc}}{RT}\right)
\end{equation}
where the penalty factor $k$ is chosen to be high enough to enforce this boundary condition. A value of $k=8\times10^5\;\text{mol}/\text{m}^2/\text{s}$ is chosen here, as it is much higher than the diffusivity, thus ensuring the boundary condition will always be fulfilled. The exponential term, with the variable $E_\text{bc}=1.71\times 10^{4}\;\text{J}/\text{mol}$, allows this penalty factor to scale as the temperature increases, such that the penalty factor always remains large compared to the diffusion and de-trapping rates. The appropriateness of the selected values of $k$ and $E_\text{bc}$ in the computational model was evaluated in \cref{subsubsection:verification_multitrap}. The use of a penalty-based boundary condition, relative to a conventional Dirichlet one ($\theta_L=0$), enables a more precise determination of the hydrogen flux, allowing for the use of coarser meshes and thus accelerating the training stage. The relation described in \cref{equation:peanlty_bc} follows that of the Tafel reaction for hydrogen desorption, in which case the constants $k$ and $E_\text{bc}$ have the physical meaning of reaction rate and energy, respectively. Hence, the kinetics of surface reactions \cite{kulagin2024kinetic,martinez2020generalised} can be accounted for in the present model. However, it should be noted that we have solely used this relation as a way to enforce a zero hydrogen boundary condition; it does not account for any surface reaction kinetics, which are assumed to be negligible, given that they occur at much faster rates than the bulk processes taking place. The effect of this boundary condition was verified by increasing and decreasing $k$ by a factor of ten, which did not alter the obtained results.

To mirror the TDS experimental process, the simulation has been divided into three sequential steps: (i) hydrogen charging of the sample, (ii) rest period, representing the transfer from the charging step to the TDS apparatus, and (iii) TDS testing at a fixed heating rate. While the charging process can be simulated within the present framework, it is common practice to charge the sample until saturation; i.e., until a uniform occupancy  $\theta_\text{L}(\textit{x,t}=0)=\theta_\text{L}^0$ is reached throughout the sample. Accordingly, this is the initial condition in our simulations. When using the Oriani model, this automatically adapts the trapped hydrogen contents to be in equilibrium with this initial concentration. In contrast, when using the McNabb-Foster model, the trapped hydrogen is initialised assuming equilibrium between the trapped and lattice concentrations, via:
\begin{equation}
    \theta_\text{T}^{(i) 0} = \frac{\theta_\text{L}K_\text{T}^{(i)}}{1+(K_\text{T}^{(i)}-1)\theta_\text{L}}
\end{equation}
with $K_\text{T}^{(i)}$ given by \cref{equation:oriani_equilibrium_constant}. 

\subsection{Numerical implementation}

Both the McNabb-Foster and Oriani frameworks were implemented in the computational model. The coupled PDEs of the McNabb-Foster formulation and the single second-order PDE of Oriani's theory are numerically solved in one-dimensional form using the finite element method, which was programmed in Python. Linear elements were used to discretise the domain, such that the hydrogen concentration is given by:
\begin{equation}
    C_\text{L} = \mathbf{N}\mathbf{c}_\text{L}
\end{equation}
with the vector $\mathbf{N}$ containing the shape functions, and the vector $\mathbf{c}_\text{L}$ the nodal values for the hydrogen concentration. As the domain is symmetric, we are only solving for half the domain. To ensure stability of the simulations, a backward Euler time discretisation scheme was used. As a result, the discretised weak form for the lattice hydrogen is given by:
\begin{equation}
\begin{split}
    & \int_\Omega \mathbf{N}^T\mathbf{N}\frac{\mathbf{c}_\text{L}^{t+\Delta t}-\mathbf{c}_\text{L}^t}{\Delta t} + D_\text{L} \nabla \mathbf{N}^T \nabla \mathbf{N}\mathbf{c}_\text{L}^{t+\Delta t} + \mathbf{N}^T J_\text{traps}\;\text{d}\Omega \\ 
    &+ \int_\Gamma \frac{k}{N_\text{L}}\exp\left( \frac{-E_\text{bc}}{RT} \right) \mathbf{N}^T \mathbf{N} \mathbf{c}_\text{L}\;\text{d}\Gamma = \mathbf{0}   
\end{split}
\end{equation}
where the interior of the domain is denoted by $\Omega$, and the edge by $\Gamma$. For the Oriani model, the trapping-related fluxes are obtained as:
\begin{equation}
\begin{split}
    J_\text{traps} =& \sum_i \frac{N_\text{T}^{(i)} K_\text{T}^{(i)}}{N_\text{L} \left[ 1 + (K_\text{T}^{(i)} - 1) \mathbf{N}\mathbf{c}_\text{L}^{t+\Delta t}/N_\text{L} \right]^2} \mathbf{N}\frac{\mathbf{c}_\text{L}^{t+\Delta t}-\mathbf{c}_\text{L}^t}{\Delta t} \\
    +& \frac{N_\text{T}^{(i)} K_\text{T}^{(i)} \Delta E^{(i)} \phi(\mathbf{N}\mathbf{c}_\text{L}^{t+\Delta t}/N_\text{L} - \left( \mathbf{N}\mathbf{c}_\text{L}^{t+\Delta t}/N_\text{L}\right)^2)}{N_\text{L} \left( 1 + (K_\text{T}^{(i)} - 1) \mathbf{N}\mathbf{c}_\text{L}^{t+\Delta t}/N_\text{L} \right)^2}
\end{split}
\end{equation}
In contrast, if using the McNabb-Foster model, the trapping-related flux is given by:
\begin{equation}
       J_\text{traps} = \frac{1}{N_\text{L}} \sum_i J_\text{trap}^{(i)}
\end{equation}
\begin{equation}
\begin{split}
    J_\text{trap}^{(i)} = N_\text{T}^{(i)} \Big( &k^{(i)} \frac{\mathbf{N}\mathbf{c}_\text{L}^{t+\Delta t}}{N_\text{L}}\left(1-\frac{\mathbf{N}\mathbf{c}_\text{T}^{(i) t+\Delta t}}{N_\text{T}}\right) \\ &- p^{(i)}\frac{\mathbf{N}\mathbf{c}_\text{T}^{(i) t+\Delta t}}{N_\text{T}} \left( \frac{\mathbf{N}\mathbf{c}_\text{L}^{t+\Delta t}}{N_\text{L}} \right) \Big)
    \end{split}
\end{equation}
Finally, for the McNabb-Foster model, this is supplemented with the discretised weak forms for the trapped hydrogen concentration:
\begin{equation}
    \int_\Omega \mathbf{N}^T\mathbf{N} \frac{\mathbf{c}_\text{T}^{(i) t+\Delta t}-\mathbf{c}_\text{T}^{(i) t}}{\Delta t} - J_\text{trap}^{(i)}\;\text{d}\Omega = \mathbf{0}
\end{equation}
These equations are solved using a Newton-Raphson-type algorithm. To ensure stability for the McNabb-Foster scheme, even for high values of the jump frequency $\nu$, a lumped integration scheme is used for the trapping terms \citep{Hageman2023a}. 

The boundary value problem is effectively one-dimensional as the TDS sample thickness $L$ is much smaller than the other dimensions. Also, symmetry conditions apply. Therefore, we consider a 1D bar of length $L/2$ with appropriate symmetry conditions, $\partial C_\text{L}/\partial x=0$ at $x=0$. A uniform mesh with 25 elements is used to discretise the sample. This discretisation is sufficient to achieve mesh-independent results within the present scheme. The time increment was selected as large as possible without losing accuracy. %The results provided in \cref{subsection:model_verification} demonstrate that the chosen time increments and mesh size are sufficient to obtain accurate results. In \cref{subsection:model_verification}, it was also verified that for high pre-exponential factors, the McNabb-Foster and Oriani models obtain the same results, and that these results match those reported in the literature, verifying the correct implementation of our model.

It should be noted that, while demonstrated with the two most widely used hydrogen transport/trapping models (McNabb-Foster, Oriani), the present ML-based framework can be applied to other modelling strategies, allowing it to encompass other physical phenomena that are not embedded in conventional hydrogen transport and trapping models, such as dislocation-mediation transport \cite{dadfarnia2015modeling,diaz2025comsol} or trap interconnectivity \cite{hoch2015effects,toribio2015generalised,wang2016influence}.

\subsection{Model verification} \label{subsection:model_verification}

A thorough validation of the developed computational model is conducted in this section. Each of the elements of the implemented formulation are verified. Firstly, the implementation of each hydrogen transport theory is assessed. The spectra obtained with both McNabb-Foster's and Oriani's models for multi-trap systems are compared with the numerical results obtained by Drexler et al. \cite{drexler_critical_2021}. Subsequently, the agreement between the McNabb-Foster and Oriani theories is assessed. Different values of trapping and de-trapping frequencies are evaluated to understand the regime of validity of Oriani's equilibrium assumption. Finally, the impact of sample frequency on the convergence of the solution is assessed.

\subsubsection{Multi-trap system} \label{subsubsection:verification_multitrap}

\begin{figure*}[t]
    \centering
    \begin{subfigure}{0.45\textwidth}
        \centering
        \includegraphics[width=\textwidth]{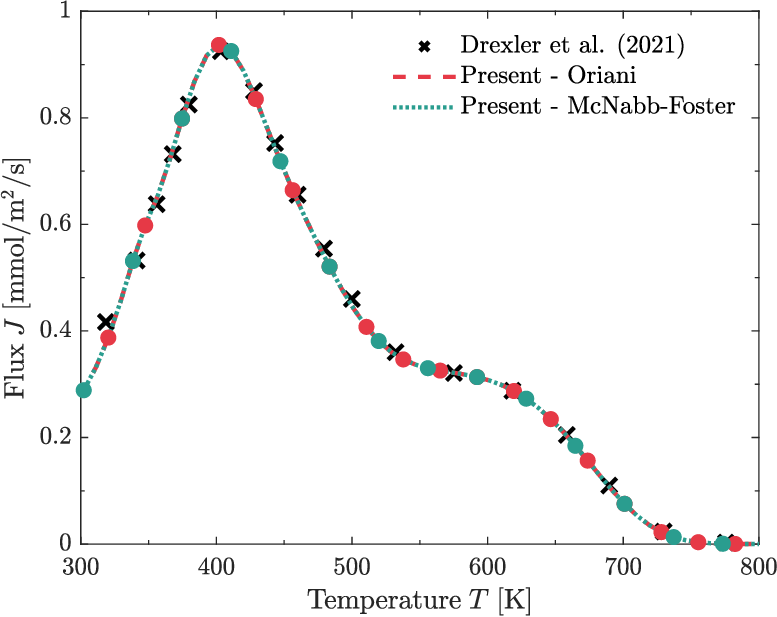}
        \caption{}
        \label{fig:validation_simulation}
    \end{subfigure}
    \hfill
    \begin{subfigure}{0.45\textwidth}
        \centering
        \includegraphics[width=\textwidth]{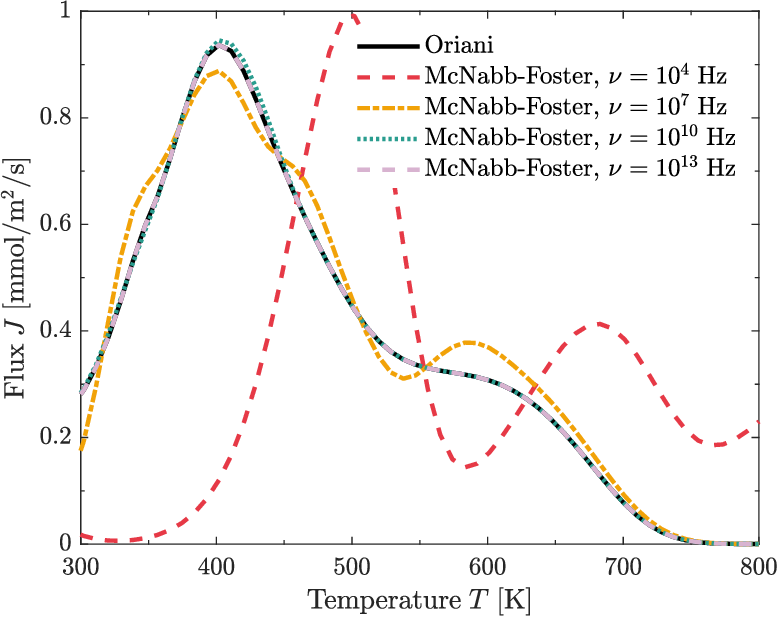}
        \caption{}
        \label{fig:validation_vibration_frequency}
    \end{subfigure}
    \caption{Validation of the Oriani and McNabb-Foster implementations for multi-trap systems: (a) comparison between the desorption spectrum reported by Drexler et al. \citep{drexler_critical_2021} and that obtained with the developed computational model, (b) comparison of TDS simulations considering the McNabb-Foster formulation for different values of jump frequency ($\nu_\text{t}=\nu_\text{d}=\nu$) and Oriani's model.}
    \label{fig:validation_simulator}
\end{figure*}

The implementation of both hydrogen transport frameworks for multi-trap systems was validated by comparing the predictions of the developed numerical model against results obtained by Drexler et al. \citep{drexler_critical_2021}, who used Oriani's method and the finite differences approach. Given that Drexler et al. \citep{drexler_critical_2021} adopted Oriani's method throughout their study, a high jump frequency, equal to the Debye frequency ($\nu=10^{13}\;\text{Hz}$), was selected for the McNabb-Foster implementation to ensure the system is within the equilibrium regime, and thus comparable to Oriani's method (see \cref{subsubsection:verification_equilibirumvalidity}). Finally, the penalty factor $k$ and variable $E_\text{BC}$ were fixed at $8\times\ 10^{5}\;\text{mol}/\text{m}^2/\text{s}$ and $1.71\times 10^{4}\;\text{J}/\text{mol}$, respectively. The results, reported in \cref{fig:validation_simulator}(a), show clear agreement between both frameworks and the reference results, confirming the correct implementation of the computational schemes and the effective enforcement of the penalty-based approach by the selected values of $k$ and $E_\text{BC}$. We have further verified that these values for $k$ are sufficiently high to enforce this boundary condition by simulating the same case with $k=k/10$, which produced identical results.

\begin{figure*}[t]
    \centering
    \begin{subfigure}{0.45\textwidth}
        \centering
        \includegraphics[width=\textwidth]{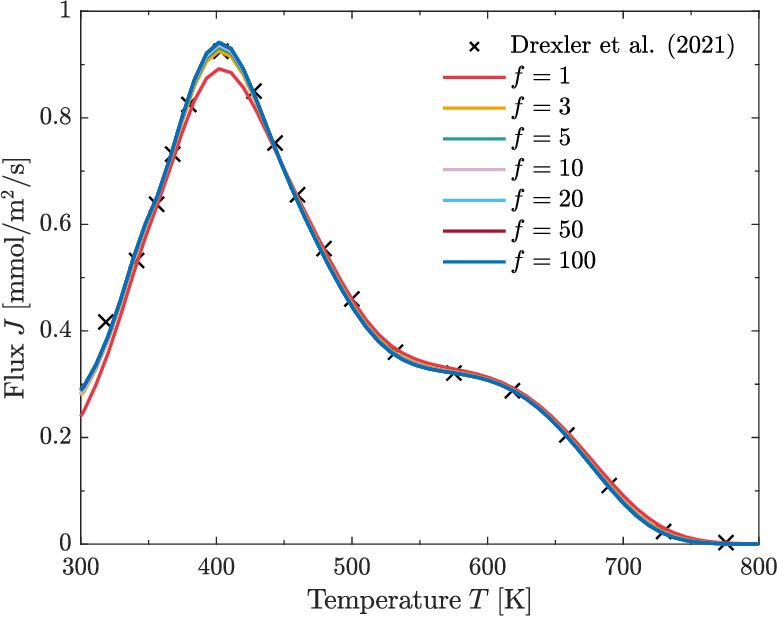}
        \caption{}
        \label{fig:McNabb_sample_frequency}
    \end{subfigure}
    \hfill
    \begin{subfigure}{0.45\textwidth}
        \centering
        \includegraphics[width=\textwidth]{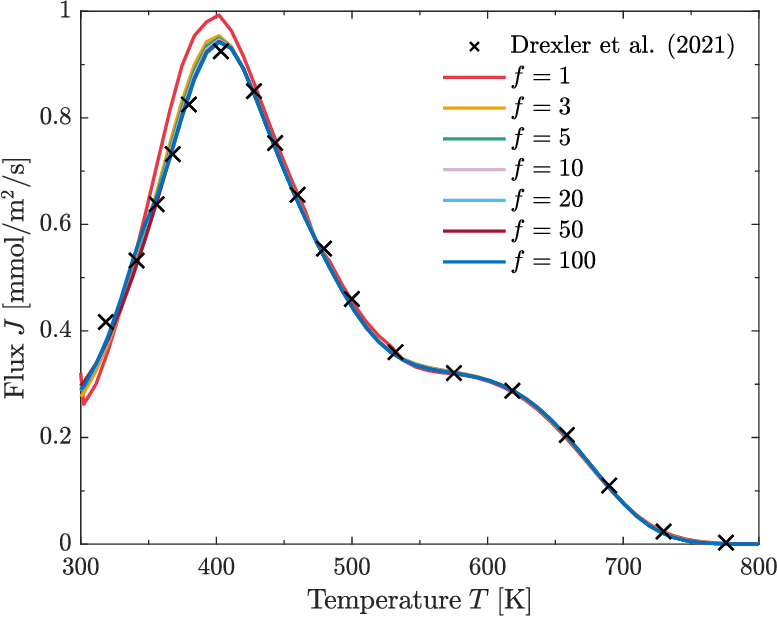}
        \caption{}
        \label{fig:Oriani_sample_frequency}
    \end{subfigure}
    \caption{Sample frequency (and time step) convergence study for (a) McNabb-Foster and (b) Oriani framework implementations. The TDS simulation predictions obtained using different sample frequencies are benchmarked against the multi-trap system simulated by Drexler et al. \citep{drexler_critical_2021}.}
    \label{fig:sample_frequency_convergence}
\end{figure*}

\subsubsection{McNabb-Foster vs Oriani models} \label{subsubsection:verification_equilibirumvalidity}

To assess the agreement between the McNabb-Foster and Oriani formulations, simulations for the same multi-trap system are carried out using these two models. Four McNabb-Foster simulations are run with increasing values of jump frequency to determine the value required for the McNabb-Foster model to converge to Oriani's model, and thus understand the validity regime of the equilibrium assumption adopted in the latter. To conduct this analysis, the multi-trap system constructed by Drexler et al. \citep{drexler_critical_2021}, previously considered in the multi-trap system validation study (\cref{subsubsection:verification_multitrap}), is selected. The results, depicted in \cref{fig:validation_simulator}(b), indicate that convergence occurs for sufficiently high values of jump frequency $\nu$. A good agreement is obtained between the Oriani and McNabb-Foster predictions for values equal to or larger than $\nu=10^{10}$, and a perfect alignment between the two models is achieved at $\nu=10^{13}$, also referred to as the Debye frequency. In the remainder of the report, the McNabb-Foster model will be implemented with a frequency that matches the Debye frequency. This choice ensures that the system remains within the equilibrium regime of validity, which will help facilitate comparisons.

\subsubsection{Time step convergence study} \label{subsubsection:verification_timestepconvergence}

A time step convergence study was conducted for both McNabb-Foster and Oriani frameworks. Simulations of the multi-trap system constructed by Drexler et al. \citep{drexler_critical_2021} were carried out for increasing values of sample frequencies, $f=1,\;3,\;5,\;10,\;20,\;50,\;100$, and compared to the original data simulated by the authors. The number of temperature evaluations that are saved as the output TDS curve was fixed at $\text{ntp}=64$, thus the time steps corresponding to the evaluated frequencies are $\Delta\text{t}=4.5,\;1.5,\;0.91,\;0.45,\;0.23,\;0.091,\;0.045\;\text{s}$.

Convergence is achieved for both McNabb-Foster and Oriani frameworks at $f=10,\; \Delta\text{t}=0.45\; \text{s}$; consequently, this value was selected for all validation studies (\cref{fig:sample_frequency_convergence}). Convergence studies were repeated for each simulated and experimental dataset, and similar results were obtained.

\subsection{Data generation} \label{subsection:data_generation}

To train our multi-NN model, datasets need to be generated based on simulated data. A dataset consists of \textit{n} data points, with each data point representing a sample. Each data point contains information on the trapping characteristics of the sample, including the number of traps, and the associated densities and binding energies. Additionally, each data point includes the temperatures and fluxes that constitute the desorption spectrum of that specific sample.

\begin{figure*}
    \centering
    \includegraphics[width=1\textwidth]{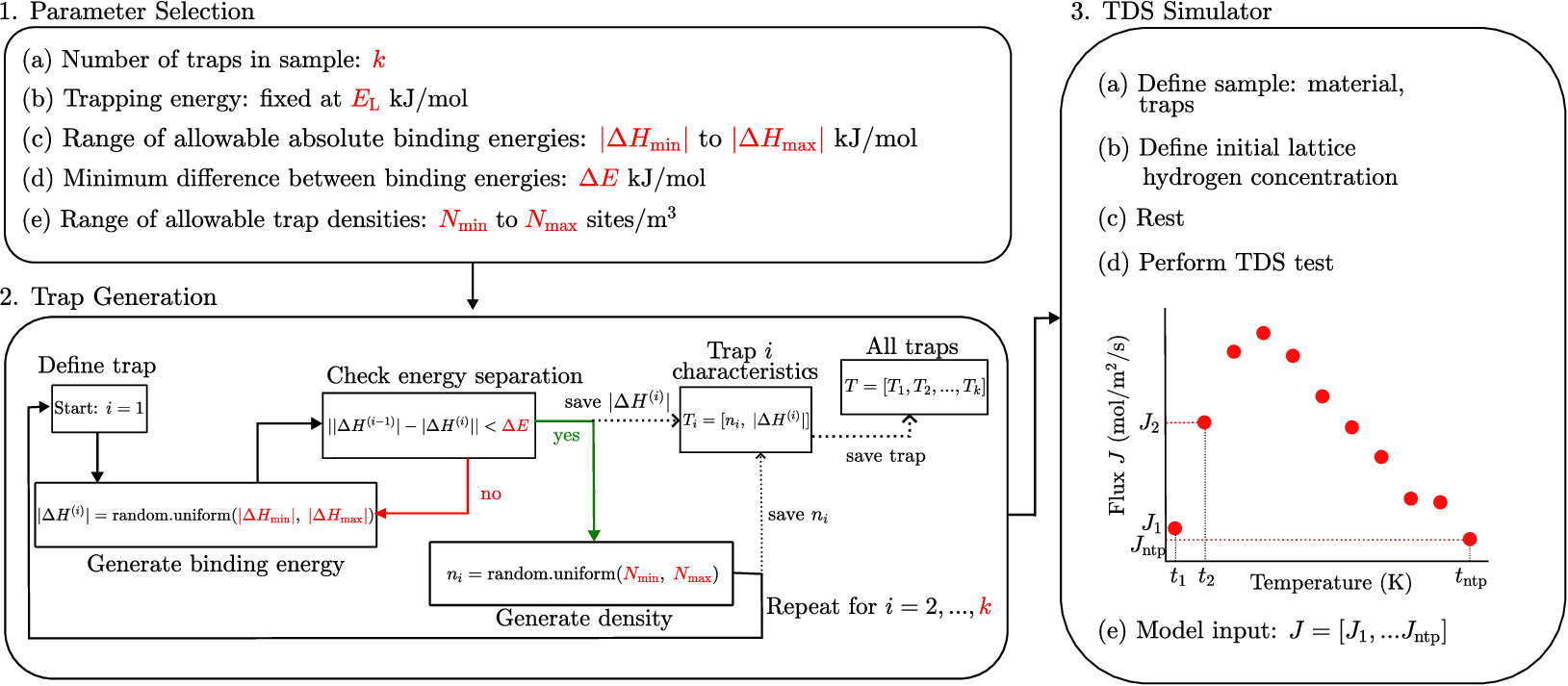}
    \caption{Schematic of the procedure used to generate a single data point corresponding to a TDS sample with \textit{k} distinct traps. The process comprises three phases: (1) selection of parameters, (2) generation of trap characteristics, and (3) TDS simulation to produce the desorption spectrum.}
    \label{fig:data_generation}
\end{figure*}

Each data point in the dataset is generated through a systematic procedure that is divided into three distinct phases. Firstly, the parameters guiding the data generation are selected. This includes the maximum number of (potential) trap types in the sample, the range of allowable trap densities, the trapping energy (fixed as the activation energy for lattice diffusion $E_\text{L}$ for all traps, as commonly done in the literature \cite{krom_hydrogen_2000}), the range of allowable absolute binding energies, and the minimum allowable separation between the binding energies of traps in the same sample. Based on these parameters, the trapping characteristics of the sample are generated. The detailed procedure is illustrated in Fig. \ref{fig:data_generation}. For each trap, the trapping energy is set to $E_\text{L}$, and the absolute binding energy is assigned a value randomly selected from the allowable range. Subsequently, the difference in the absolute binding energy of the newly generated trap and the previously generated traps in the sample is evaluated to ensure all traps are slightly distinct. If the difference is less than the minimum allowable separation, the binding energy generation step is repeated until a suitable value is found. The trap is then assigned a density, randomly chosen from the permissible range, and its characteristics are stored alongside those of the other traps. To ensure each TDS spectrum corresponds to a unique set of trapping energies, these traps are ordered from lowest to highest energy. This uniqueness is required for the ANN to be able to relate TDS curves to trap characteristics.

A critical step in the first two phases of the data generation process is defining the range of allowable trap densities and binding energies that encompass all possible trap configurations that could contribute to the TDS spectrum under consideration. Trap densities and absolute binding energies can vary significantly. Shallow traps can have absolute binding energies $|\Delta H|<30\;\text{kJ}/\text{mol}$ while the binding energies of deep traps can be $|\Delta H|> 60\;\text{kJ}/\text{mol}$. Large differences in $N_\text{T}$ and $|\Delta H|$ make it challenging to cover the entire trap domain effectively using the random generation approach. Thus, for each test case, it is essential to identify the regime of interest \textemdash based on observed flux magnitudes and peak temperatures of the TDS spectrum\textemdash to simplify the data generation process and ensure that only relevant spectra are considered for model training. 

In the final phase, the simulation of the TDS experiment is carried out. This requires several additional material, test, and numerical parameters that characterise the experimental TDS procedure and material used. The material parameters consist of the pre-exponential diffusion coefficient $D_0$, initial lattice hydrogen concentration $C_0$, density of interstitial lattice sites $N_\text{L}$, and activation energy for lattice diffusion $E_\text{L}$. If the McNabb-Foster framework is employed, the vibration frequency is also required. This is set equal to the Debye frequency ($\nu=10^{13}\;\text{Hz}$), as this is its expected magnitude \citep{garcia-macias_tds_2024}. The required test parameters are: sample thickness $L$, rest time $t_\text{rest}$, heating rate $\phi$, and minimum $T_\text{min}$ and maximum $T_\text{max}$ temperatures. For the numerical parameter inputs, one only needs the number of temperature evaluations ($ntp$), which defines how many desorption flux values are recorded, and the sampling frequency ($f$), which specifies the number of time steps between each recorded value. The TDS test time $t_\text{test}$ can be estimated using $t_\text{test}=(T_\text{max}-T_\text{min})/\phi$. The time increment can then be calculated using $\Delta t=t_\text{test} / (ntp \times f )$. The frequency enables us to reduce the number of temperature evaluations that are saved without increasing the time increment, ensuring stability is achieved for all cases while allowing the number of inputs for the ANN to still be reasonable.

\section{Machine learning approach} \label{section:ML_approach}

\begin{figure}
    \centering
    \includegraphics[width=0.7\linewidth]{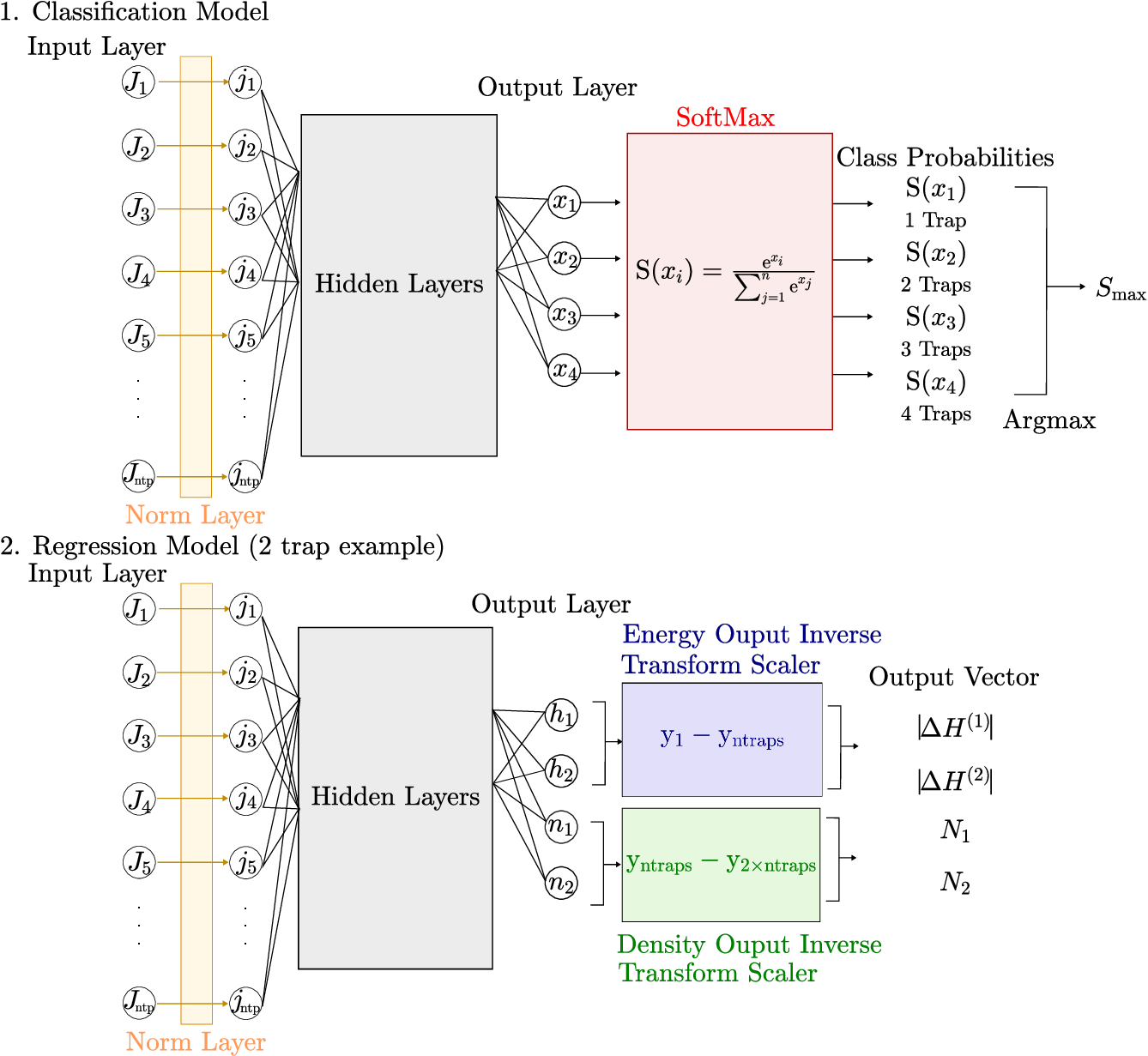}
    \caption{Overview of the NN-coupled TDS-based analysis framework for characterising hydrogen trapping parameters. Two multi-layer, fully connected, feed-forward NNs are employed. The first NN, a classification model, extracts the number of traps present in the sample. The second NN, a regression model, predicts the binding energies and densities of each trap. The variables in the regression model’s output vector, $e_i$ and $n_i$, represent the scaled absolute binding energy and density of trap $i$, respectively. These scaled parameters are transformed back to their original units and scale, $|\Delta H^{(i)}|$ and $N_\text{T}^{i}$, by applying the appropriate inverse scaler.}
    \label{fig:flowchart_model_predictions}
\end{figure}

The multi-NN model proposed for the quantitative analysis of TDS spectra comprises two multi-layer, fully connected, feed-forward NNs trained with back-propagation \citep{aggarwal_introduction_2023, aggarwal_deep_2023, theodoridis_neural_2020}. The detailed structure and flowchart of the multi-NN model are illustrated in \cref{fig:flowchart_model_predictions}. The first step involves passing the input data to the first NN, referred to as the classification model, which predicts the number of traps present in the material. An intermediate stage follows, in which the appropriate second NN, the regression model, is selected. This model is trained to predict trapping properties based on the number of traps, allowing the appropriate model to be chosen based on the classification model's prediction. Finally, the input data is fed into the regression model, which computes the densities and absolute binding energies. Both neural network models are implemented using the Keras machine-learning framework \citep{Tensorflow, Keras}. In this section, we will first describe the optimised structure of the ANN models, after which we will briefly discuss the parameter and re/post-processing procedure tuning that has been performed to reach this optimised architecture.

\subsection{Optimised multi-model neural network architecture}

The input vector for both neural networks consists of hydrogen desorption rate data from the thermal desorption spectra; specifically, $64$ flux values sampled at evenly spaced temperature intervals. To enhance convergence and stability, a minimum threshold of $10^{-10}\;\text{mol}/ (\text{m}^{2} \, \text{s})$ is applied to all flux values. Additionally, due to the significant differences in magnitude among the fluxes, a log transformation is applied to the input data, resulting in the following calculation for each flux value:
\begin{equation}
    J_\text{input}^n = \text{log}\left(\max\left<J^n,10^{-10}\right>\right)
\end{equation}
No normalisation is performed beforehand, as the Keras-provided normalisation layer is added after the input layer in both networks, eliminating the need for pre-processing of this type. This layer applies z-score standardisation, scaling the (log-scaled) input fluxes based on the average and standard deviation of the training dataset. The effect of this scaling is demonstrated in \cref{sec:modelDataPreprocessing}.

\subsubsection{Hidden layers}
As the input vector size is $64$, the input layer for both NNs comprises $64$ neurons. These inputs are passed onto the different layers, which differ between the NNs. The hidden layer topology of the regression model can be summarised as 64-64-32-16-8, all multiplied by the number of outputs, e.g., 256-256-128-64-32 for predicting the energies and densities for a two-trap system. The choice to scale the number of nodes in the hidden layers with the number of outputs of the NNs was made to account for the increase in complexity associated with the increase in the number of traps. In the classification model, the hidden layers are structured as 256-128-64-32. The rectified linear (ReLU) activation function is used throughout the layers \citep{theodoridis_neural_2020}.

\subsubsection{Post-processing}
The two models also differ in their post-processing procedures. In the classification model, the softmax activation function is used in the output layer to convert the output values into discrete probabilities (see \cref{fig:flowchart_model_predictions}). The maximum probability is then taken as the prediction for the number of trapping sites. In the regression model, the output vector is divided into two vectors: values corresponding to trap energies (positions 1 to $\textit{ntp}$) and densities ($\textit{ntp}+1$ to $2\times\textit{ntp}$). Each vector is passed through its respective inverse transform scaler to recover the actual (unscaled) absolute binding energy and density values, which are then recombined into a single output vector. Separate scalers are used for absolute binding energies and densities due to their significantly different magnitudes\textemdash energies are typically 2 to 3 orders of magnitude greater than densities. When a single combined scaler was used, it was observed that the larger energy values dominated the scaling process, adversely affecting the learning of density values. This imbalance led to significantly higher prediction errors in the density outputs, justifying the use of two independent scalers. 

\subsubsection{Model training}

Training is achieved by employing the supervised learning process, which intends to map input examples of TDS data to a known number of traps and densities/energies for the classification and regression models. Prior to training the model, the training datasets must be generated. In addition to the data generation parameters specified in \cref{subsection:data_generation}, we also define the maximum number of traps that the multi-NN model is expected to predict. The total number of datasets generated corresponds to this maximum value. For instance, if the maximum number of traps is set to 4, four distinct datasets are created. Each dataset contains data points representing TDS samples with a distinct number of traps, ranging from 1 in the first dataset to the maximum (4) in the final dataset. The data points of each dataset are generated following the procedure outlined in \cref{fig:data_generation}, with the number of traps $k$ set to the specific number of traps for that dataset. Each dataset is subsequently used to train a regression model corresponding to the number of traps of the respective dataset. Finally, the classification model is trained using the combined set of all datasets.

All ML models were trained with a dataset comprising 50,000 data points (i.e., simulations). To evaluate the NN model's performance during training, the database of simulation results was split into training and validation sets, using an 80/20 split. An additional test set comprising 500 data points was generated to evaluate the final model. Given that the training data is synthetic and thus unlimited datasets can be generated, the simple hold-out validation approach is adopted, which is appropriate for this purpose \citep{aggarwal_deep_2023}.

The training process uses the same optimiser for both models, Adamax, due to its stability and efficiency in training deep networks, with learning rate and weight decay set to $10^{-3}$ \citep{theodoridis_neural_2020}. The choice of loss function, however, is task-dependent \citep{aggarwal_deep_2023, santosh_deep_2022}. For the regression model, we use Mean Squared Error (MSE):
\begin{equation} \label{equation:mse} \text{MSE}=\frac{1}{n}\sum_{i=1}^{n}(j_{i}-j_{i}')^2 \end{equation}
where $n$ is the number of predictions, $J_{i}$ is the true output (i.e., simulation results), and $J_{i}'$ is the predicted output from the neural network. MSE is well-suited for regression-type problems because it heavily penalises larger errors, encouraging precise, continuous-value predictions \citep{malitckii_evaluation_2020}. However, MSE is inappropriate for classification tasks, as it assumes a continuous output and fails to model class probabilities effectively. For classification, we instead use cross-entropy loss:
\begin{equation} \label{equation:cross-entropy} \text{cross-entropy loss}=-\sum_{i=1}^n \sum_{k=1}^{m}  j_{ik}\ln{j'_{ik}}\end{equation}
where $m$ is the number of output nodes in the neural network. When the number of output layer nodes matches the number of classes (as in our model), cross-entropy corresponds to the negative log-likelihood of the observations. Thus, the loss function penalises confident but incorrect predictions, making it ideal for classification tasks by encouraging accurate probability distributions \citep{theodoridis_neural_2020}. The batch size was fixed at $32$ for both models. The number of training epochs was set to $100$ and $200$, multiplied by the number of outputs, for the classification and regression models, respectively.

\subsection{Model optimisation: hyperparameter tuning}

The hyperparameter tuning process consists of identifying the combination of hyperparameters that optimises the efficiency, performance, and generalisation of the models \citep{aggarwal_deep_2023}. This ensures that it provides the best possible predictions given a fixed amount of training data, reducing the computational cost of both the data generation and the training process. The optimal set of hyperparameters is highly dependent on factors such as the nature of the inputs (i.e. data distribution, dimensionality and quality), the outputs, and the complexity of the task. Since these factors vary significantly across the regression and classification ANNs, the hyperparameter tuning process was repeated for each model individually to account for these differences. 

For each NN, the hyperparameter tuning process followed a systematic approach, divided into three sequential stages. Each stage focused on optimising a different category of hyperparameters: architecture, algorithm, and regularisation hyperparameters. During the entire tuning process, several hyperparameters were kept constant: the number of training epochs ($100\times\text{number of outputs}$), the batch size (32), and the number of training data points (10,000). The hyperparameter tuning is performed on a lower amount of training data compared to the actual model training, to allow a large range of hyperparameter combinations to be explored. However, it is expected that the parameters optimised here will still be (close to) optimised for larger amounts of training data.

\subsubsection{Model architecture}

The first stage involved optimising the model architecture by tuning key hyperparameters\textemdash specifically, the number of hidden layers and the number of nodes per layer. Recommendations exist in literature on how to approach this process \cite{haque_ann_2002, haque_prediction_2001, xu_artificial_2006}. In this work, the network topology was chosen by systematically varying the architecture while monitoring model performance. For the regression models, we evaluated three levels of complexities (i.e., 128-64, 64-64 and 128-128) defined by the number of nodes in the first two hidden layers. Subsequent layers decreased progressively in size; for instance, in a 128-64 topology, the third layer has 32 nodes, the fourth layer 16 nodes, and so forth. The number of nodes in each hidden layer is scaled by a factor proportional to the number of model outputs, allowing complexity to increase with output dimensionality. For each complexity, topologies of four to seven hidden layers were tested. A similar analysis was conducted for the classification model, with five levels of complexity (i.e., 128-64, 256-128, 512-256, 256-256, 128-128) and topologies ranging from three to five hidden layers. The scaling factor in this case was the maximum number of output classes (i.e., maximum potential traps). Throughout the analysis, certain hyperparameters were held constant: the Adamax optimiser was used with a learning rate and weight decay of $10^{-3}$, and ReLU was applied as the activation function in all hidden layers. 

\begin{figure*}[t]

    \centering
    \text{Classification Model}
    \vspace{0.2cm}
    
    \begin{minipage}{0.45\textwidth}
        \centering
        \begin{subfigure}[b]{\textwidth}
            \includegraphics[width=\textwidth]{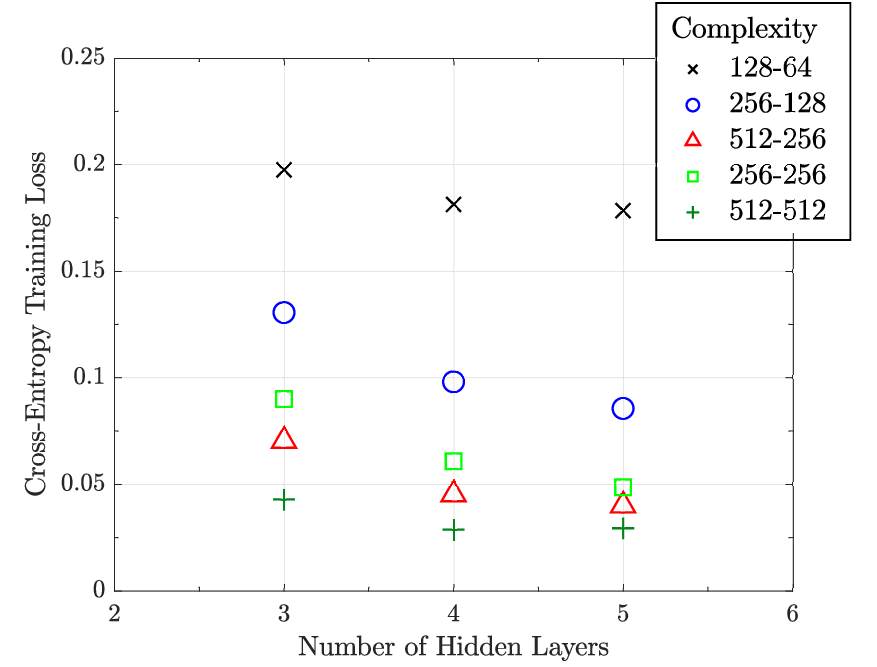}
            \caption{}
            \label{fig:topology_class}
        \end{subfigure}
    \end{minipage} 
    \begin{minipage}{0.45\textwidth}
        \centering
        \begin{subfigure}[b]{\textwidth}
            \includegraphics[width=\textwidth]{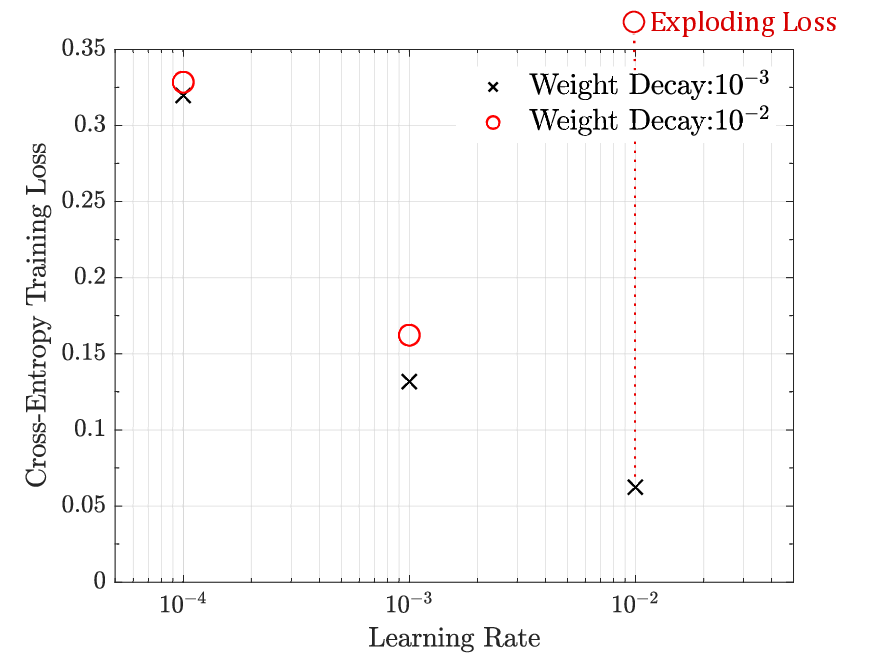}
            \caption{}
            \label{fig:lr_wd_class}
        \end{subfigure}
    \end{minipage}
    
    \centering
    \text{Regression Model}
    \vspace{0.2cm}
    
    \begin{minipage}{0.45\textwidth}
        \centering
        \begin{subfigure}[b]{\textwidth}
            \includegraphics[width=\textwidth]{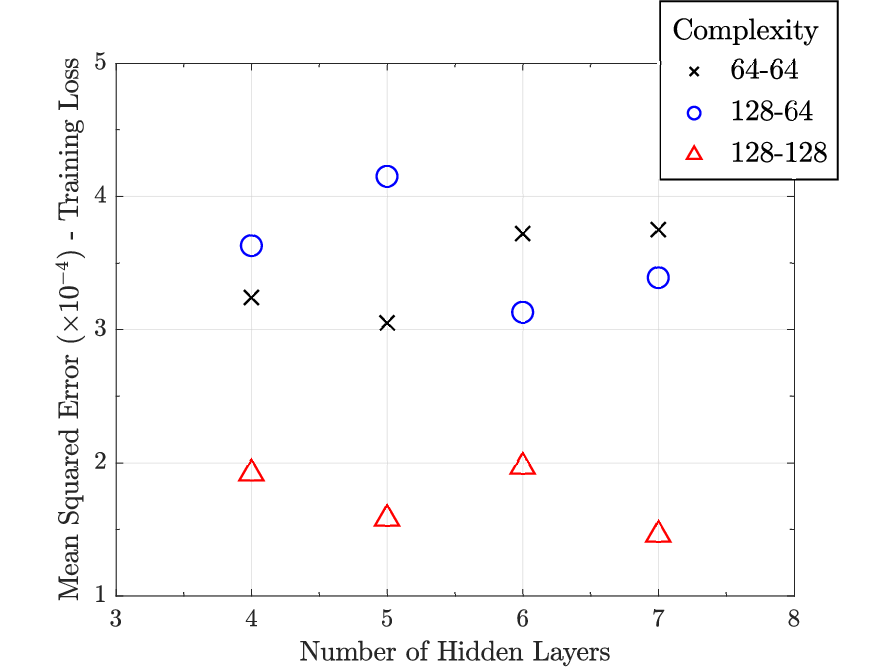}
            \caption{}
            \label{fig:topology_reg}
        \end{subfigure}
    \end{minipage} 
    \begin{minipage}{0.45\textwidth}
        \centering
        \begin{subfigure}[b]{\textwidth}
            \includegraphics[width=\textwidth]{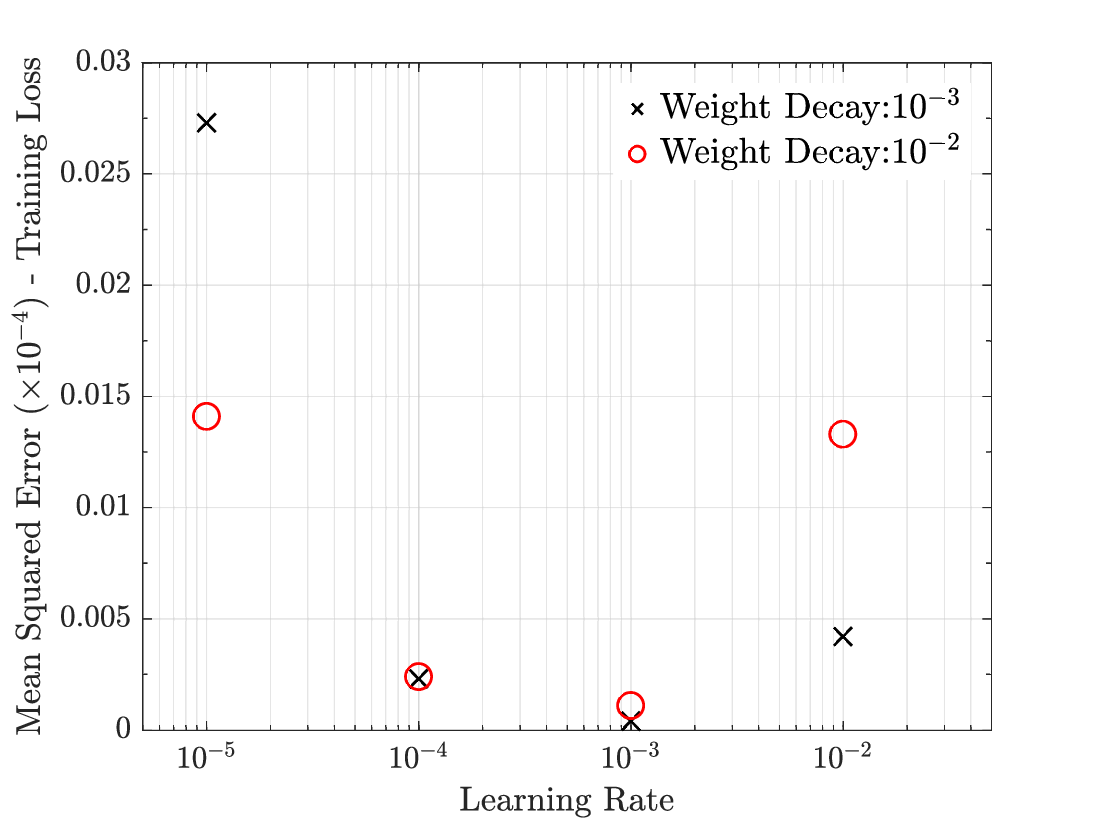}
            \caption{}
            \label{fig:lr_wd_reg}
        \end{subfigure}
    \end{minipage}
    \caption{Hyperparameter tuning results for (a, b) classification and (c, d) regression models. Subfigures (a) and (c) show the effect of the number of hidden layers on training loss for various model complexities. Complexity is defined by the number of nodes in the first two hidden layers, with subsequent layers halving in size (e.g., a 128-64 topology continues with 32, then 16 nodes). Subfigures (b) and (d) show how training loss varies with the learning rate (Adamax optimiser), evaluated at two weight decay values: $10^{-2}$ and $10^{-3}$.}
    \label{fig:hp_tuning}
\end{figure*}

The results are shown in \cref{fig:topology_reg} and \cref{fig:topology_class} for regression and classification models, respectively. For the regression model, increasing the number of layers or nodes did not significantly reduce the training loss, which remained between $1.5\times10^{-4}$ to $4.5\times10^{-4}$. Thus, a simple topology comprising five hidden layers (i.e., 64-64-32-16-8) was chosen to minimise computational cost. In contrast, the classification model benefited from increased complexity. For instance, in three-layer topologies, increasing the number of nodes in the first two hidden layers from 128–64 to 512–256 reduced the training loss by more than 0.1 (see \cref{fig:topology_class}). However, the benefit of adding more layers diminishes: for each level of complexity, the training loss with five layers is nearly identical to that with four (see \cref{fig:topology_class}). A topology of medium complexity comprising 4 hidden layers (i.e., 256-128-64-32) was selected. More complex topologies were not considered, as mitigating over-fitting in such models would require significantly more training data, increasing both data generation and model training computational costs.

\subsubsection{Algorithmic hyperparameters}

With regards to the algorithm hyperparameter stage, different combinations of learning rate and weight decay were investigated. Keras Tuner \citep{Keras} was used to conduct a grid search tuning study. The values of learning rate and weight decay investigated were $[10^{-5}, 10^{-4}, 10^{-3}, 10^{-2}]$ and $[10^{-3}, 10^{-2}]$ respectively. The optimised topologies were selected to conduct the study. The results obtained for the classification and regression models are reported in \cref{fig:lr_wd_class} and \cref{fig:lr_wd_reg}, respectively. For both NNs, a learning rate and weight decay of $10^{-3}$ is required to achieve optimal performance. 

\subsubsection{Regularization}  

A preliminary study was performed to evaluate the suitability of the most commonly used regularisation techniques for mitigating over-fitting, specifically penalty-based regularisation and dropout layers. Dropout layers were incorporated after each hidden layer, with dropout rates of 0.2, 0.3, and 0.5 tested. For penalty-based regularisation, L2 regularisation was assessed with values ranging from $10^{-3}$ to $10^{-1}$, applied exclusively to the hidden layers. The inclusion of these techniques did reduce over-fitting; however, it also led to a degradation in model performance. This trade-off renders regularisation ineffective in improving overall model performance. Consequently, regularisation techniques were not integrated into the final model. The most effective approach to mitigating over-fitting identified in this study was to increase the number of training data points. 

\subsection{Model optimisation: data pre-processing}
\label{sec:modelDataPreprocessing}

A major challenge in applying ANN to TDS curves arises from the fact that both input and output parameters span several orders of magnitude. Conventional ML schemes assume the inputs and outputs to be linearly spaced, as is the case for the trapping energy. These energies are typically between $10-100\;\text{kJ}/\text{mol}$ \citep{li_review_2020, drexler_model-based_2019}, with errors in the estimates of these energies being equal throughout this range, e.g. a difference of $\pm5 \;\text{kJ}/\text{mol}$ has the same effect whether it is $50\pm5$ or $150\pm5\;\text{kJ}/\text{mol}$. This energy relates to the horizontal location of the peaks within a TDS curve in a (mostly) linear manner, making it a fairly easy quantity for ANN to predict. In contrast, trapping densities span many orders of magnitude, e.g., ranging from $10^{23}$ to $10^{28}\;\text{sites}/\text{m}^3$ \citep{fernandez-sousa_analysis_2020, yen_critical_2003}. Using these densities directly in the NN would result in bias, as the network would favour accurately predicting the higher densities, which contribute to the largest mean squared error (MSE). To ensure the prediction error scales with the predicted quantity, a pre-processing algorithm is applied to transform the density into a linear scale. 

To achieve this, different combinations of transformation and scaling techniques were applied to the input data features and output predictions with the objective of adjusting the distributions to more closely resemble a normal distribution. For the input data (the TDS curves), Log and Yeo-Johnson power transformations \citep{Yeo2000} were used to reduce the skewness and kurtosis of each feature's distribution. Both transformations achieved similar effects; however, the log transformation was selected due to its lower computational cost.

Following the transformation, normalisation techniques were applied. Both min-max scaling and z-score scaling were assessed after applying the log transformation. Models trained with log-transformed and z-score-scaled data achieved a lower mean absolute error (MAE), leading to the selection of z-score scaling as the optimal normalisation technique. While normalisation techniques do not impact the skewness or kurtosis, the mean and standard deviation were checked to confirm successful scaling, with values approximately equal to 0 and 1 for all features.

\begin{figure*}[t]
    \centering
        \begin{subfigure}{0.32\textwidth}
            \centering
            \includegraphics[width=\textwidth]{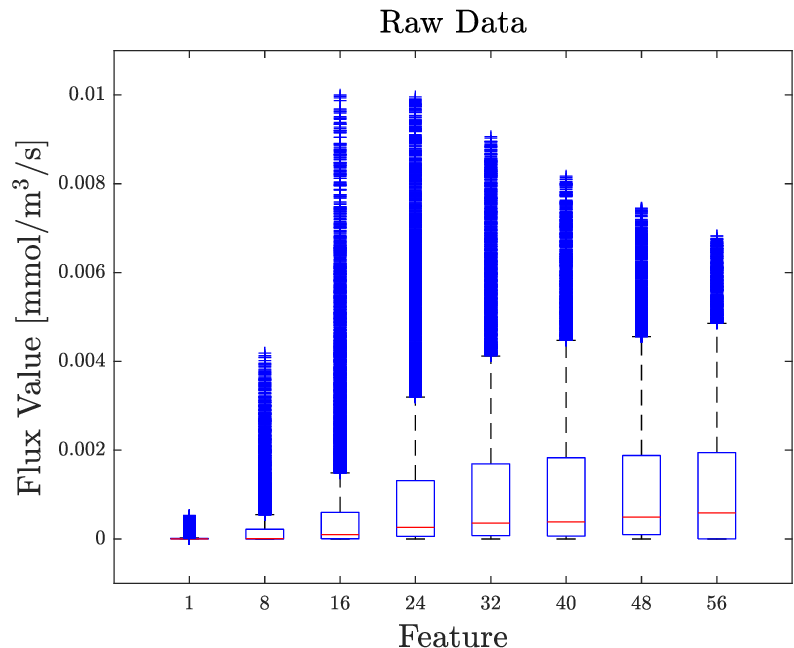}
            \caption{}
            \label{fig:boxplots_raw}
        \end{subfigure}
        \hfill
        \begin{subfigure}{0.32\textwidth}
            \centering
            \includegraphics[width=\textwidth]{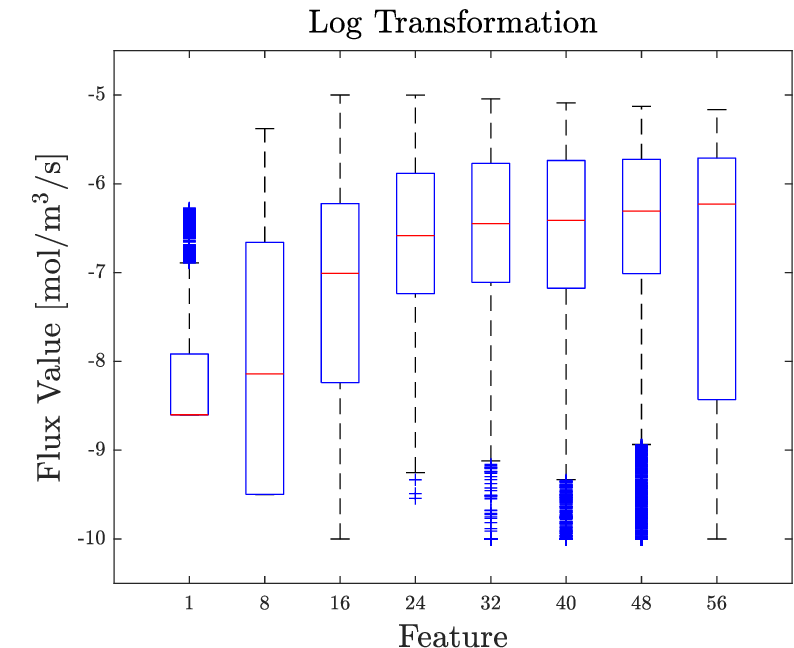}
            \caption{}
            \label{fig:boxplots_log}
        \end{subfigure}
        \hfill
        \begin{subfigure}{0.32\textwidth}
            \centering
            \includegraphics[width=\textwidth]{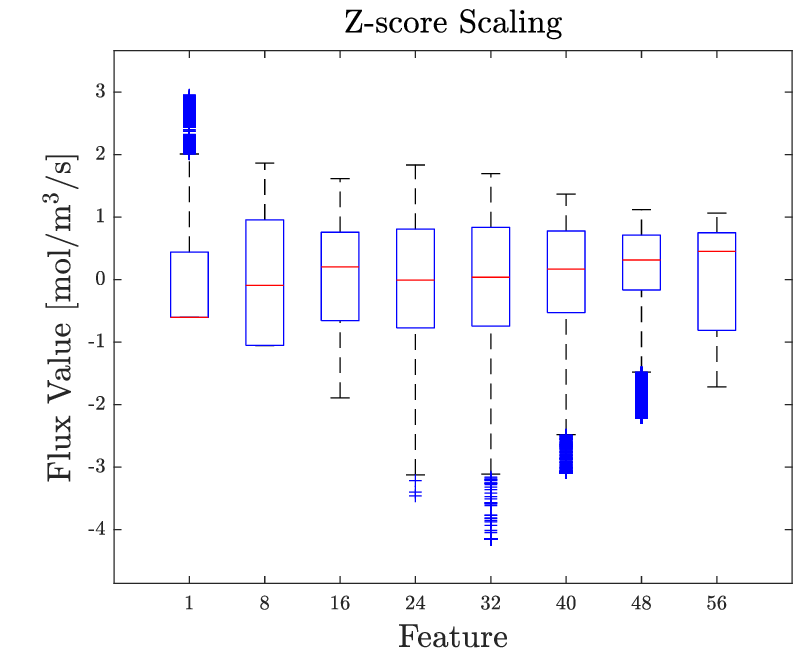}
            \caption{}
            \label{fig:boxplots_log_zscore}
        \end{subfigure}
    \caption{Box-plots showing the distribution of eight selected features at each pre-processing stage: (a) raw data, (b) log-transformed data, and (c) log-transformed data with z-score scaling. Each feature represents the desorption flux value corresponding to a distinct temperature of the TDS spectrum. The selected features are evenly distributed across the temperature range, representing the overall feature set. The pre-processing steps progressively reduce skewness and standardise the data, aiming to approximate a normal distribution with zero mean and comparable variance across all features.}
    \label{fig:feature_distributions}
\end{figure*}

The impact of each stage of the optimised pre-processing technique on the distribution of eight input features is depicted in \cref{fig:feature_distributions}. The selected features are evenly distributed across the temperature range, representing the entire feature set. The raw data is highly skewed and deviates significantly from a normal distribution (\cref{fig:boxplots_raw}). After applying the log transformation (\cref{fig:boxplots_log}), there is a significant reduction in skewness. Notably, the skewness of feature 8, which initially exhibited the highest skewness, decreased from 3.4 to 0.2. With the addition of z-score scaling (\cref{fig:boxplots_log_zscore}), the distribution of all features more closely resembles a normal distribution, having a mean and a standard deviation approximately equal to 0 and 1, respectively. Despite the transformations, several outliers remain, likely due to the highly skewed nature of the raw data, as observed in \cref{fig:boxplots_log_zscore}. To further improve the distributions, a threshold of $10^{-10}$ was applied to all flux values to clip very small values before applying the transformation and normalisation techniques.

Finally, zero-mean Gaussian noise was introduced into the training data. This step is essential for accurately fitting experimental data, as the synthetic training data is noise-free, while real-world data is inherently noisy. The first case study was used to determine the appropriate level of noise, with three noise levels tested \textemdash corresponding to standard deviations of 0.01, 0.05, and 0.1, representing increasing intensity. The optimal noise level was identified as that corresponding to a standard deviation of 0.05. The analysis was repeated for the second and third case studies to verify consistency across scenarios. 

\section{Parameter identification of experimental TDS data} 
\label{sec:results}

To demonstrate the ability of ML models to determine the trapping characteristics\textemdash number of trapping defects, trap binding energies and trap densities\textemdash from experimental TDS spectra, three test cases were analysed corresponding to: (1) a high-strength AISI 4340 tempered martensitic steel; (2) a tempered martensitic Fe-C-Ti alloy, containing 0.1 wt\% C and a stoichiometric amount of Ti; and (3) a tempered martensitic Fe-0.05C-0.20Ti-2.0Ni alloy. 
For each test case, the experimental TDS spectra were fitted using the NN procedure outlined in Section \ref{section:ML_approach}. To evaluate the accuracy and reliability of the developed ML models, an analysis of each TDS spectrum was also carried out using an already established approach\textemdash applying the inference fitting capabilities embedded in the \texttt{TDS Simulator} App described by Garcia-Macias et al. \citep{garcia-macias_tds_2024}. This fitting is based on the particle swarm optimisation (PSO) algorithm implemented in MATLAB. Throughout the remainder of the paper, we refer to the fitting of experimental data using 
the \texttt{TDS Simulator} App \citep{garcia-macias_tds_2024} as the `conventional optimisation' approach, which serves as a reference to assess the accuracy of our machine-learning-based approach against conventional optimisation schemes. For each test case, the test and material parameters were kept consistent across both the conventional and the ML analyses to ensure that the results were directly comparable.

\subsection{High-strength AISI 4340 tempered martensitic steel} \label{section:Novak_experimental_data_fit} 

The first test case involves three TDS curves obtained by Novak et al. \citep{novak_statistical_2010} for a high-strength AISI 4340 tempered martensitic steel, with each curve corresponding to a different controlled heating rate: $200$, $100$, and $50^{\circ}\text{C}/\text{h}$. The conventional optimisation and ML analyses were conducted separately for each TDS spectrum. The material and data generation parameters were kept constant across all analyses. Conversely, the test parameters differed by one variable, the heating rate, which necessitated the use of different ML models for each case. As a result, three distinct multi-NN models were trained, each corresponding to one of the TDS spectra.

Test parameters were selected based on the values reported by the authors. The sample thickness was set to $L = 0.0063\;\text{m}$, as specified, with the minimum and maximum temperatures set to $T_\text{min} = 293.15\;\text{K}$ ($20^{\circ}\text{C}$) and $T_\text{max} = 873.15\;\text{K}$ ($600^{\circ}\text{C}$). The resting time was not explicitly stated. Here, we assume this to be $45\;\text{min}$ ($t_\text{rest} = 2700\;\text{s}$), a typical resting time. As noted by Raina et al. \citep{raina_analysis_2018}, the resting time can significantly influence the desorption spectrum, particularly the initial flux values. Hence, this introduces a potential source of error between our predictions and the actual trapping parameters. Material parameters were taken from the properties characteristic of the bcc lattice, see \cref{table:Material_Properties}, as the material in question is a tempered martensitic steel. This table also includes the molar mass, $M_\text{M}$, and the density $\rho_\text{M}$, which are required to convert the TDS hydrogen fluxes between the reported units of $\text{wppm}$, and the units of $\text{mol}/\text{m}^3$ used throughout our work. 

\begin{table*}[h!]
    \centering
    \begin{tabular}{c c c c c c c}
    \toprule
    Metal/alloy & $\textit{E}_\text{L}\;[\text{kJ/mol}]$ & $\textit{D}_\textit{0}\;[\text{m}^2/\text{s}]$ & $\textit{M}_\text{M}\;[\text{g}/\text{mol}]$ & $\rho_\text{M}\;[\text{g}/\text{cm}^{3}]$ & $\textit{N}_\text{L}\;[\text{sites}/\text{m}^3]$ & $\textit{C}_\text{L}^{0}\;[\text{mol}/\text{m}^3]$ \\
    \midrule
    \text{Bcc lattice} & 5690 & $7.23\times10^{-8}$ & 55.847 & 7.847 & $5.1\times10^{29}$ & 0.06 \\
    \bottomrule
    \end{tabular}
    \caption{Typical values of relevant material parameters for a bcc lattice. Data taken from \cite{garcia-macias_tds_2024}.}
    \label{table:Material_Properties}
\end{table*}

The data generation parameters were chosen by identifying a realistic range of trapping binding energies and densities that would produce fluxes matching the magnitudes of the observed TDS spectra. The range for absolute binding energies was set between $50$ and $150\;\text{kJ/mol}$, with a minimum difference between binding energies of $10\;\text{kJ/mol}$. Trap densities were constrained to a range of $0.1-10\;\text{mol}/\text{m}^3$ ($6.022\times10^{22}-6.022\times10^{24}\;\text{sites/m}^{3}$). The ML models were trained with $50,000$ data points to predict a maximum of $4$ traps. For the conventional optimisation analysis, the default settings in \texttt{TDS Simulator} were employed for the numerical simulation parameters (200 temperature evaluations and 100 discretisation elements) and for the global optimisation algorithm, as per the referenced study \cite{garcia-macias_tds_2024}.

\begin{figure*}[t]
    \centering
    \begin{subfigure}{0.32\textwidth}
        \centering
        \includegraphics[width=\linewidth]{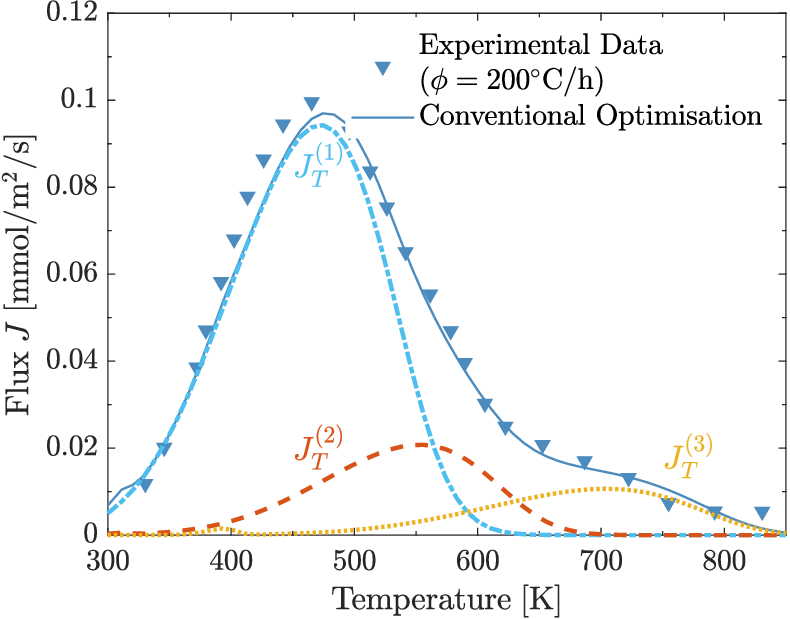}
        \caption{}
        \label{fig:Novak_1_Conventional}
    \end{subfigure}
    \hfill
    \begin{subfigure}{0.32\textwidth}
        \centering
        \includegraphics[width=\linewidth]{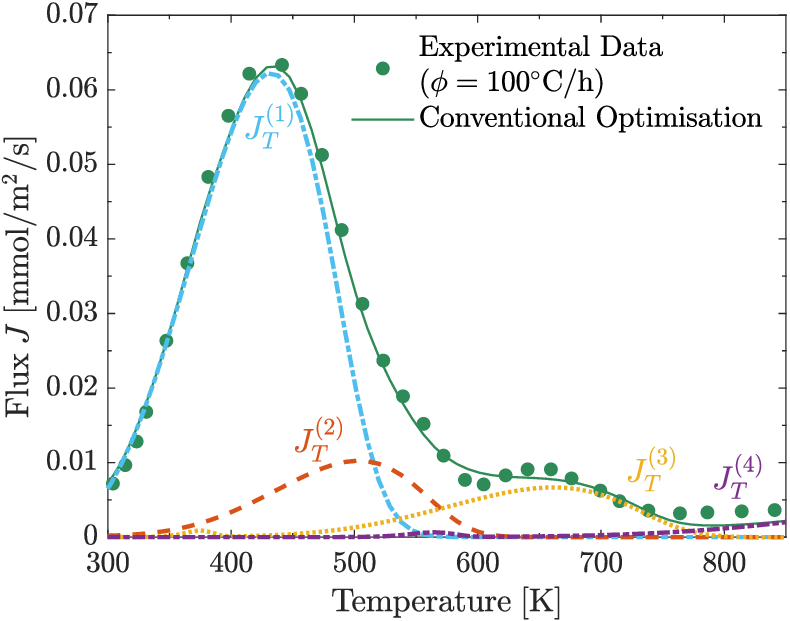}
        \caption{}
        \label{fig:Novak_2_Conventional}
    \end{subfigure}
    \hfill
    \begin{subfigure}{0.32\textwidth}
        \centering
        \includegraphics[width=\linewidth]{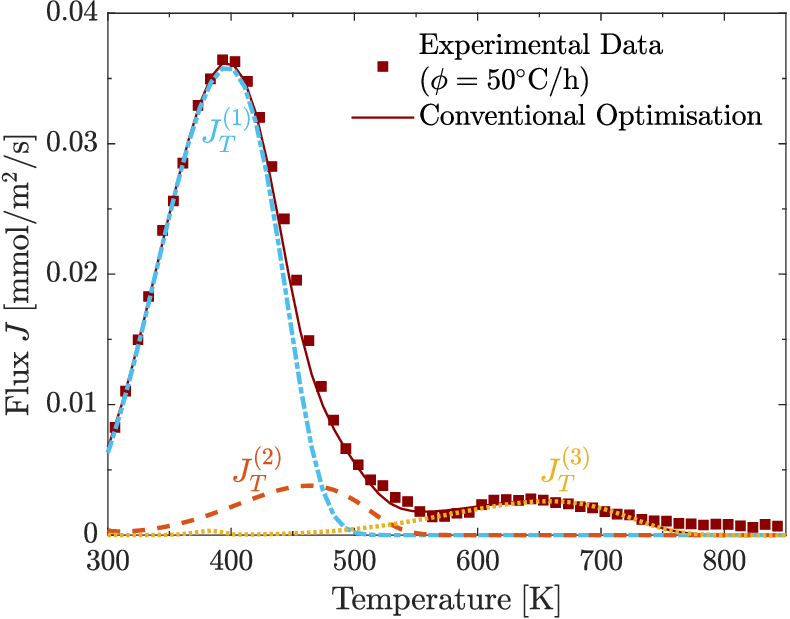}
        \caption{}
        \label{fig:Novak_3_Conventional}
    \end{subfigure}
    \caption{Using conventional optimisation (\texttt{TDS Simulator} \citep{garcia-macias_tds_2024}) to gain insight into the trapping characteristics of a high-strength tempered martensitic steel. The experimental \citep{novak_statistical_2010} and simulated desorption curves, with the latter being based on the trapping parameters fitted by \texttt{TDS Simulator}'s optimisation algorithm, are presented for different temperature ramps: (a) $\phi=200^\circ\text{C/h}$, (b) $\phi=100^\circ\text{C/h}$, and (c) $\phi=50^\circ\text{C/h}$. The contribution of each relevant trap type $i$, denoted $J_T^{(i)}$, as determined by \texttt{TDS Simulator}, is also depicted. The traps are reported in order of absolute binding energy $|\Delta H|$.} \label{fig:Novak_Conventional}
\end{figure*}

The results of the conventional optimisation analysis for all heating rates are presented in \cref{fig:Novak_Conventional}, showing both the simulated desorption curves and the individual trap contributions. The optimisation algorithm is able to find TDS spectra that accurately follow the experimental data, successfully capturing the major features. Across all heating rates, the underlying trapping behaviour remained consistent, as reflected in the repeated identification of three dominant traps. These exhibited similar binding energies: Trap 1 ranged from $-47$ to $-54\;\text{kJ/mol}$ (blue peak), Trap 2 from $-64$ to $-72\;\text{kJ/mol}$ (orange peak), and Trap 3 from $-94$ to $-98\;\text{kJ/mol}$ (yellow peak). While slight variations in absolute trap densities were observed, their relative contributions to the dominant peak remained stable. A fourth, high-energy, low-density trap (purple peak) appeared only at the intermediate heating rate of $\phi=100^\circ\text{C/h}$.

\begin{figure}
    \centering
    \includegraphics[width=0.5\textwidth]{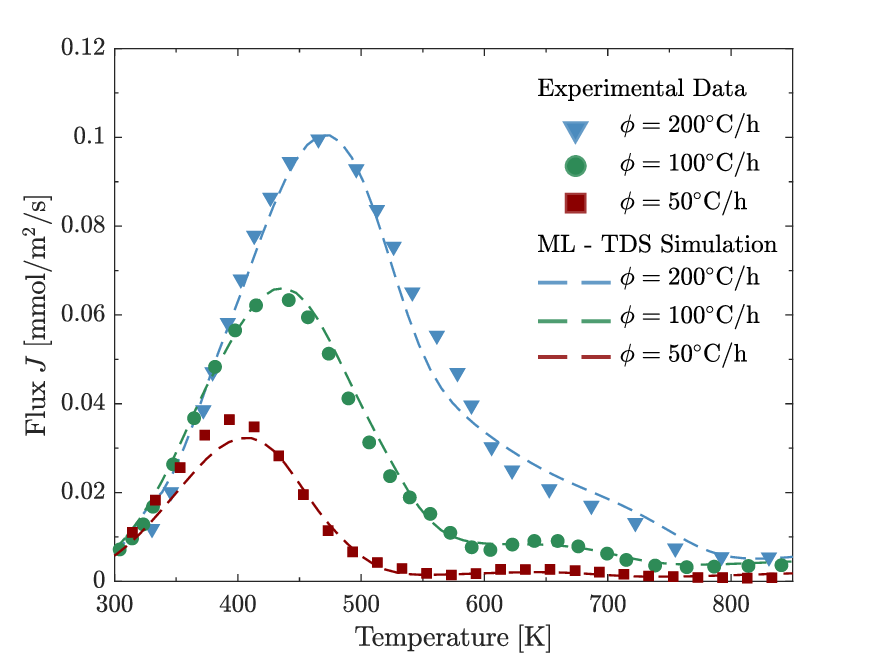}
    \caption{Using our ML approach to gain insight into the trapping characteristics of a high-strength tempered martensitic steel. The experimental \citep{novak_statistical_2010} and simulated desorption curves, with the latter being based on the trapping parameters fitted by the ML models, are presented for different temperature ramps: $\phi=200^\circ\text{C/h}$, $\phi=100^\circ\text{C/h}$, and $\phi=50^\circ\text{C/h}$. }
    \label{fig:Novak_ML_All}
\end{figure}

\begin{figure*}[t]
    \centering
    \begin{subfigure}{0.32\textwidth}
        \centering
        \includegraphics[width=\linewidth]{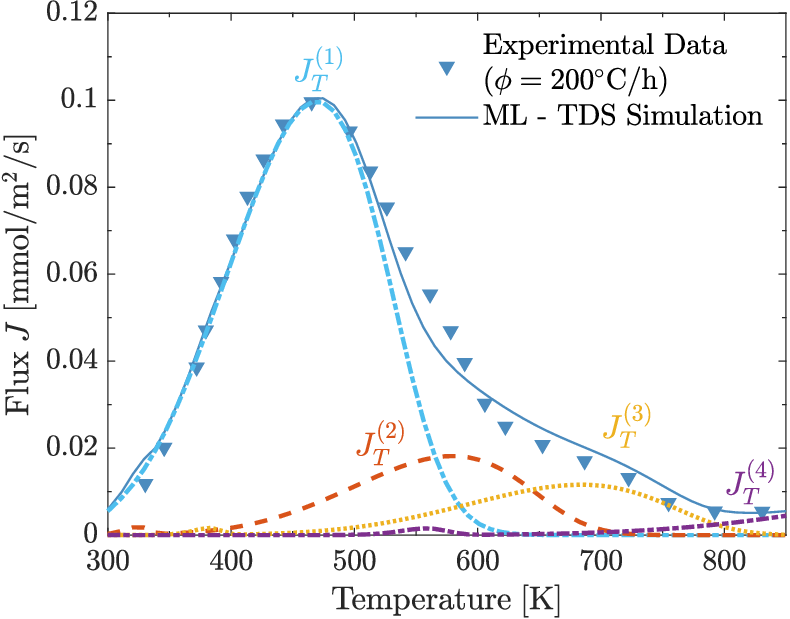}
        \caption{}
        \label{fig:ML_Novak_1}
    \end{subfigure}
    \hfill
    \begin{subfigure}{0.32\textwidth}
        \centering
        \includegraphics[width=\linewidth]{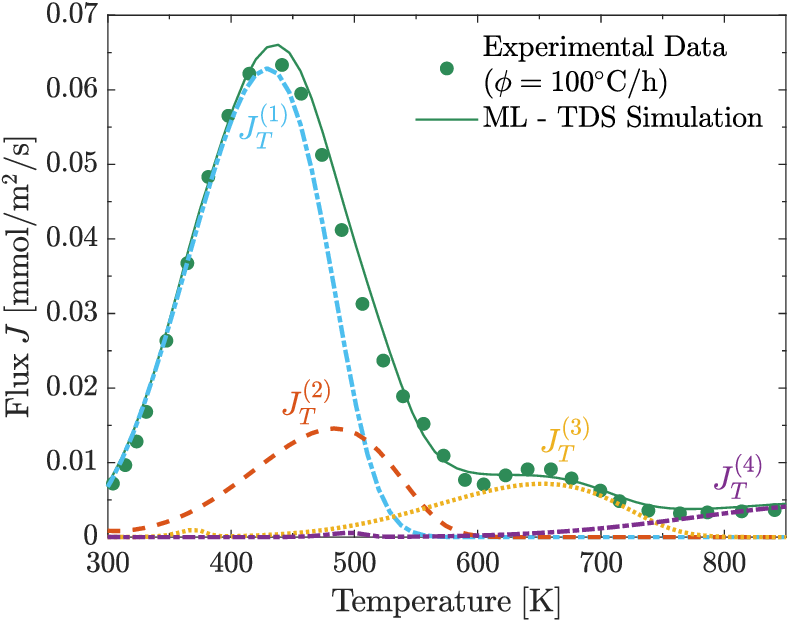}
        \caption{}
        \label{fig:ML_Novak_2}
    \end{subfigure}
    \hfill
    \begin{subfigure}{0.32\textwidth}
        \centering
        \includegraphics[width=\linewidth]{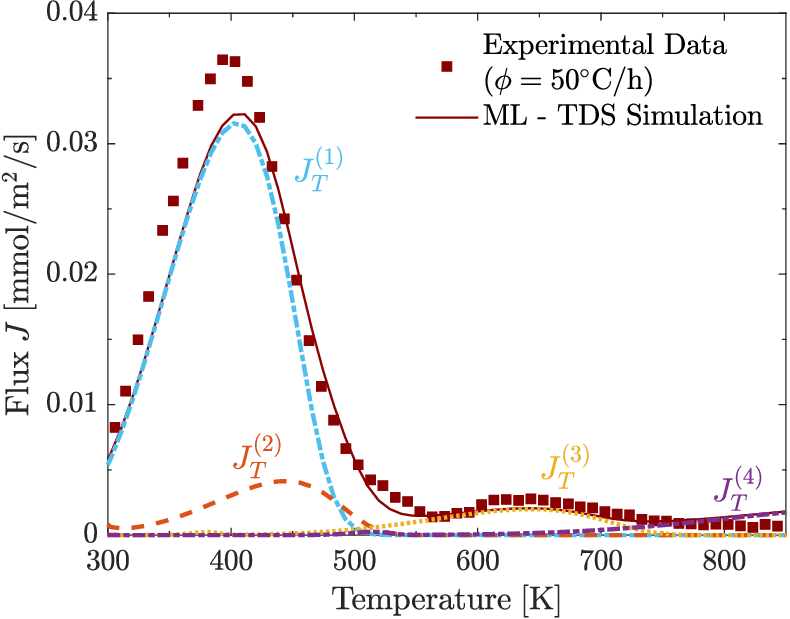}
        \caption{}
        \label{fig:ML_Novak_3}
    \end{subfigure}
    \caption{Contribution of each relevant trap type $i$ to the hydrogen desorption spectrum, denoted $J_T^{(i)}$, for different temperature ramps: (a) $200^{\circ}\text{C}/\text{h}$, (b) $100^{\circ}\text{C}/\text{h}$, and (c) $50^{\circ}\text{C}/\text{h}$, as determined by the ML approach. The traps are reported in order of absolute binding energy $|\Delta H|$.}
    \label{fig:Novak_ML_Predictions}
\end{figure*}

The results obtained using the ML approach are shown in \cref{fig:Novak_ML_All}. For all heating rates, the ML predictions closely replicate the main features of the experimental spectra, capturing both the prominent low-temperature peaks and the smaller, higher-temperature peaks. Some minor quantitative discrepancies are noted at the lowest heating rate ($\phi=50^\circ\text{C/h}$), particularly in the prominent peak height. The individual trap contributions identified by the ML model are shown in \cref{fig:Novak_ML_Predictions}, again indicating consistent trapping behaviour across all heating rates. Specifically, the peak temperature corresponding to each trap type is almost identical across all spectra. Additionally, the order of trap type densities is the same ($\text{Trap 1}>\text{Trap 2}>\text{Trap 3}>\text{Trap 4}$). The inferred binding energies are similar to those obtained from the conventional optimisation approach: Trap 1 ranged from $-48$ to $-53\;\text{kJ/mol}$, Trap 2 from $-61$ to $-75\;\text{kJ/mol}$, and Trap 3 from $-90$ to $-96\;\text{kJ/mol}$. An additional trap (Trap 4) with binding energy ranging from $-128$ to $-145\;\text{kJ/mol}$ was identified for all heating rates.

The trap type corresponding to each identified peak can be determined by comparing the values of $\Delta H$ obtained in this study with similar ranges of values reported for various tempered martensitic steels. Several trap types can be attributed to peak 2, in particular those related to incoherent interfaces, such as prior-austenite grain boundaries \citep{li_hydrogen_2004, novak_statistical_2010}. The relatively high $|\Delta H|$ reported for peak 3 can instead be associated with only a few trap states, specifically incoherent carbides and carbon matrix interfaces \citep{li_hydrogen_2004}. Determining the trap type associated with peak 1 is less straightforward. Employing Kissinger's approach, Novak et al. \citep{novak_statistical_2010} reported peak 1 to be associated with the elastic strain field of dislocations (or other martensitic interfaces), having $\Delta H = -18\; \text{kJ}/\text{mol}$. In this study, Trap 1 was found to have a value of $\Delta H$ ranging from $-47$ to $-54\;\text{kJ}/\text{mol}$. Several studies have reported dislocations in tempered martensitic steels to have a $\Delta H = -25\;\text{to}\;-35\;\text{kJ}/\text{mol}$ \citep{fernandez-sousa_analysis_2020, li_review_2020}. Given the underestimation in (absolute) binding energies associated with Kissinger's approach, it is reasonable to attribute Trap 1 to the presence of dislocations. Finally, peak 4 is most likely a consequence of experimental artefact. The trapping contribution is very small, and no trap sites are reported in literature having such large values of $|\Delta H|$.

\begin{figure*}[t]
    \centering
    \includegraphics[width=\textwidth]{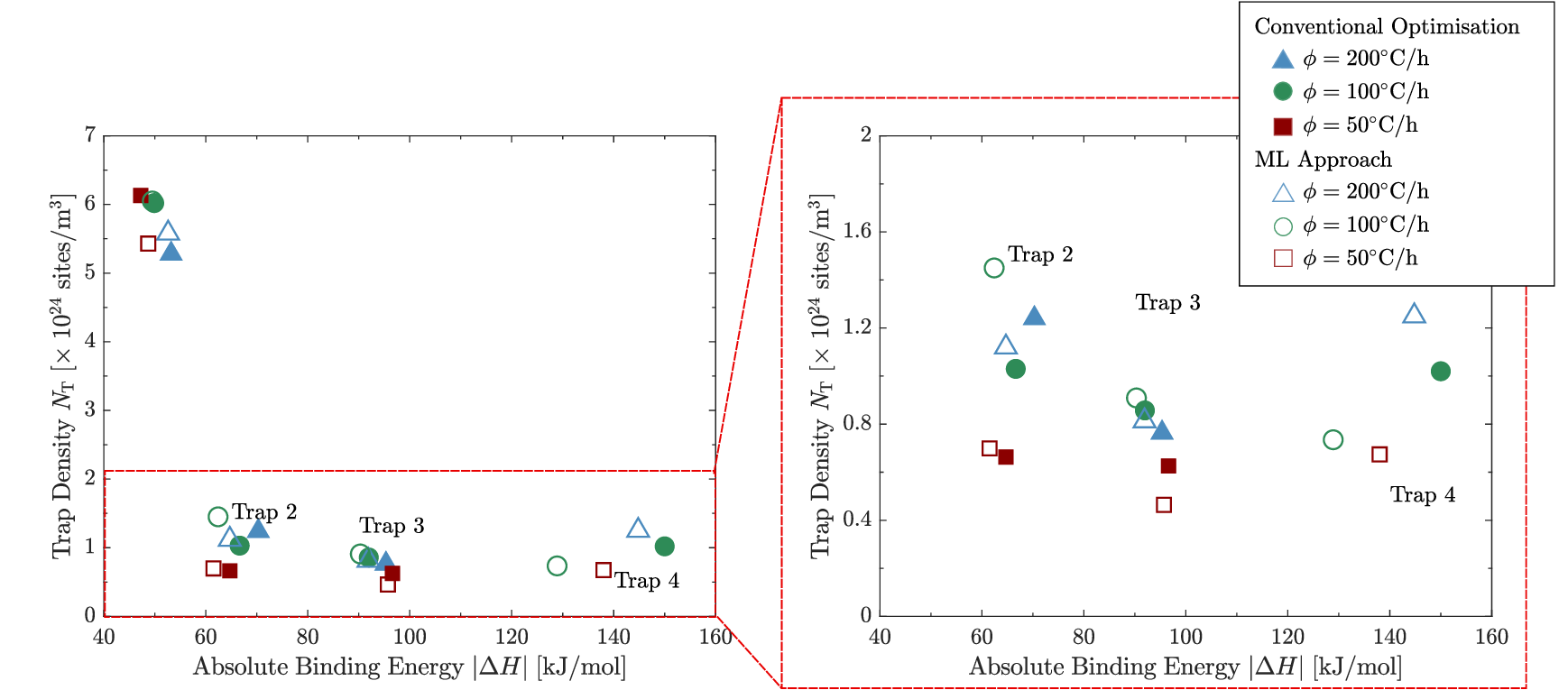}
    \caption{Scatter plot comparing the absolute binding energies ($|\Delta  H|$) and trap densities ($N_\text{T}$) obtained using conventional optimisation (\texttt{TDS Simulator} \citep{garcia-macias_tds_2024}) and ML approaches. The plot illustrates the agreement between the two approaches in predicting the trapping characteristics of a high-strength tempered martensitic steel from experimental TDS spectra for different temperature ramps: $200^{\circ}\text{C}/\text{h}$, $100^{\circ}\text{C}/\text{h}$, $50^{\circ}\text{C}/\text{h}$. The traps are reported in order of absolute binding energy $|\Delta H|$.}
    \label{fig:Novak_Comparison_Plot_Trapping_Parameters}
\end{figure*}

A direct comparison of the trapping parameters\textemdash number of traps, binding energies, and densities\textemdash obtained through both conventional optimisation and ML approaches is presented in \cref{fig:Novak_Comparison_Plot_Trapping_Parameters}. The comparison highlights the quantitative differences between the two methods, particularly in terms of trap densities. For each heating rate, the inferred densities of each trap type differ slightly between the approaches.
A prominent example is the trap densities of Trap 2 and Trap 3 for $\phi = 100^\circ \text{C}/\text{h}$. Conventional optimisation predicts densities of $N_\text{T}^{(2)}=1.0 \times 10^{24}\;\text{sites}/\text{m}^3$ and $N_\text{T}^{(3)}=8.6 \times 10^{23}\;\text{sites}/\text{m}^3$, while the ML approach predicts densities of $N_\text{T}^{(2)}=1.5 \times 10^{24}\;\text{sites}/\text{m}^3$ and $N_\text{T}^{(3)}=9.1 \times 10^{23}\;\text{sites}/\text{m}^3$. Differences are also evident for Trap 4, in both trap density and binding energy. Conventional optimisation predicts a density of $N_\text{T}^{(4)}=1.0 \times 10^{24}\;\text{sites}/\text{m}^3$ and a binding energy of $\Delta H^{(4)}=-150.0\;\text{kJ/mol}$, while the ML approach predicts a density of $N_\text{T}^{(4)}=7.4 \times 10^{23}\;\text{sites}/\text{m}^3$ and a binding energy of $\Delta H^{(4)}=-128.9\;\text{kJ/mol}$. The larger discrepancies for Trap 4, especially in binding energy, likely arise from the fact that this peak was not fully captured in the spectrum, which complicates the analysis and enhances the ill-posed nature of the problem.

\begin{figure*}[t]
    \centering
    \begin{subfigure}{0.32\textwidth}
        \centering
        \includegraphics[width=\linewidth]{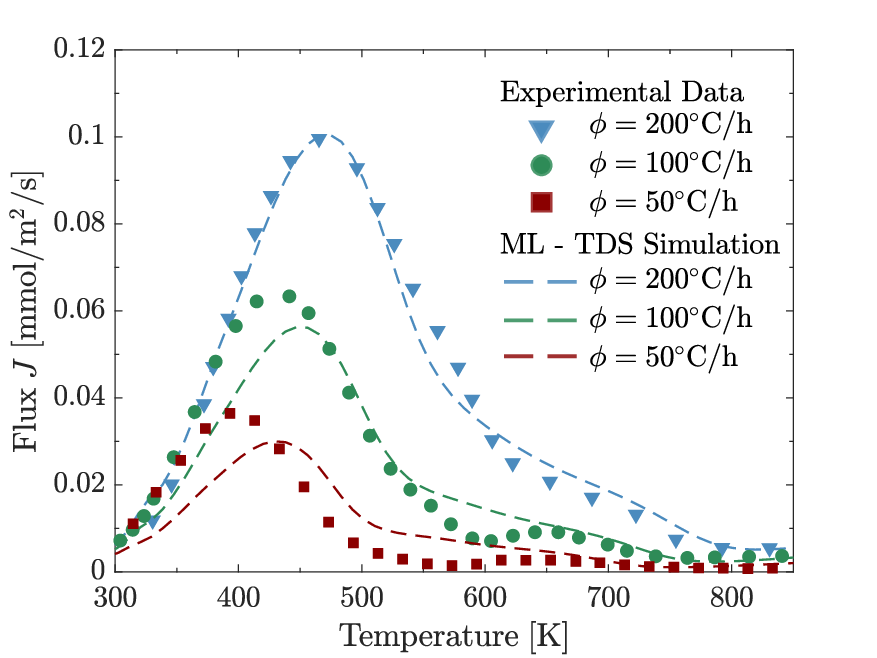}
        \caption{}
        \label{fig:Novak_1_ML_HR_Comparison}
    \end{subfigure}
    \hfill
    \begin{subfigure}{0.32\textwidth}
        \centering
        \includegraphics[width=\linewidth]{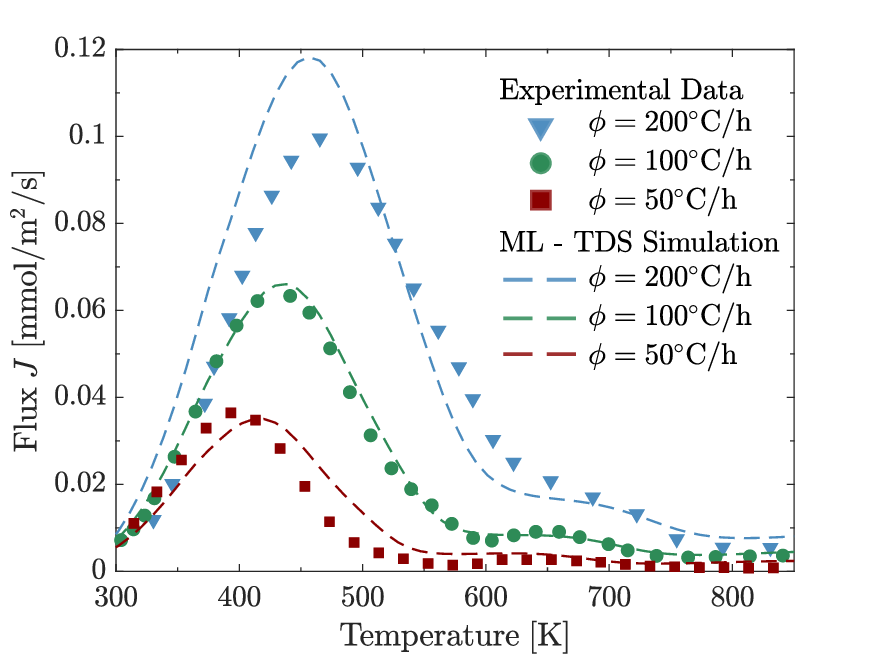}
        \caption{}
        \label{fig:Novak_2_ML_HR_Comparison}
    \end{subfigure}
    \hfill
    \begin{subfigure}{0.32\textwidth}
        \centering
        \includegraphics[width=\linewidth]{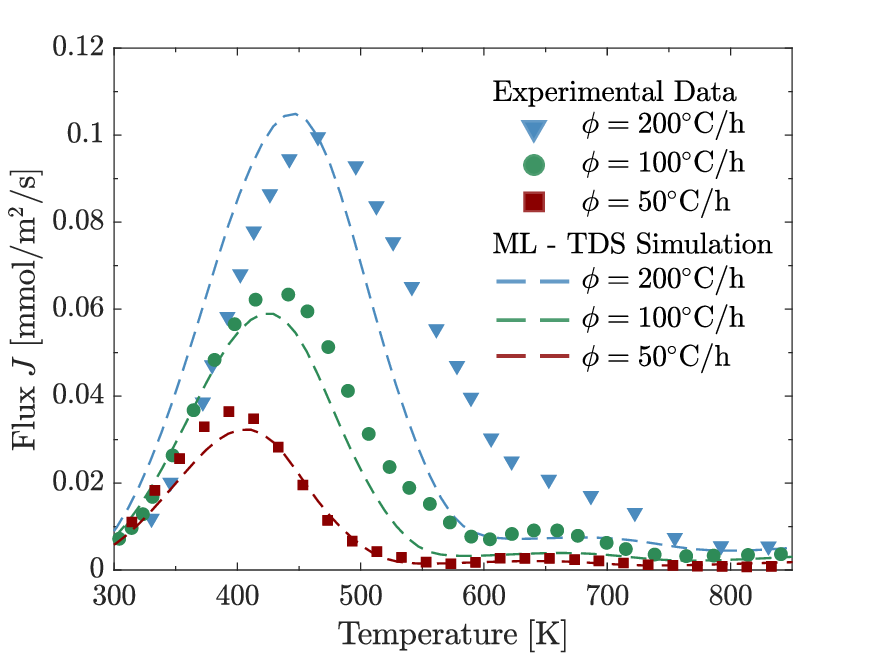}
        \caption{}
        \label{fig:Novak_3_ML_HR_Comparison}
    \end{subfigure}
    \caption{Comparison of simulated and experimental TDS data of high-strength tempered martensitic steel \citep{novak_statistical_2010} at various heating rates. Simulated desorption curves use trapping parameters predicted by the ML models for experimental spectra obtained with temperature ramps of: (a) $200^{\circ}\text{C}/\text{h}$, (b) $100^{\circ}\text{C}/\text{h}$, and (c) $50^{\circ}\text{C}/\text{h}$.}
    \label{fig:Novak_ML_HR_comparison}
\end{figure*}

\cref{fig:Novak_Comparison_Plot_Trapping_Parameters} illustrates that while both the ML and conventional optimisation approaches are able to reproduce the spectra for different heating rates, the results for one heating rate cannot be extrapolated to other heating rates. As the same material was used for all tests, identical trapping parameters are expected across varying heating rates; however, this is not observed. For instance, the ML predictions show that Trap 3 for $\phi = 50^\circ\text{C}$ has a density of $\Delta H^{(3)}=4.6 \times 10^{23}\;\text{sites}/\text{m}^3$, significantly lower than that of the other two heating rates: $\Delta H^{(3)}=8.1 \times 10^{23}\;\text{sites}/\text{m}^3$ at $\phi = 200^\circ\text{C}$ and $\Delta H^{(3)}=9.1 \times 10^{23}\;\text{sites}/\text{m}^3$ at $\phi = 100^\circ\text{C}$. Additionally, there is a noticeable difference in the binding energy of Trap 2, with a disparity of over $10\;\text{kJ/mol}$ between $\phi = 200^\circ\text{C}$ and the other two heating rates. These quantitative differences are highlighted when the parameters identified for each heating rate are used to generate the spectra of the other heating rates (\cref{fig:Novak_ML_HR_comparison}). Mismatches in terms of both peak location and height can be observed between the simulated spectra and the experimental data. This mismatch has been observed in several studies, especially at higher heating rates \citep{garcia-macias_tds_2024,hurley_numerical_2015,simoni_integrated_2021}. It has been argued that these discrepancies are due to deviations of the experimental temperature ramp from the assumed linear ramp used in theoretical models. Such deviations may arise from laboratory setups that cannot ensure a perfectly linear or reproducible temperature ramp. Since key material parameters (such as $D_\text{L}$, $k^{(i)}$ and $p^{(i)}$) are governed by Arrhenius relationships, minor temperature differences can have a significant effect. There is also an inevitable sample-to-sample variability, and the deviation from the experimental data could well fall within the experimental scatter. Nevertheless, both sample-to-sample variability and temperature profile errors are becoming smaller in modern TDS systems.

\subsection{Tempered martensitic Fe-C-Ti alloy}
\label{Sec:CaseStudy2}

For the second example, the ML approach is employed to determine the trapping characteristics of a Fe-C-Ti alloy, containing 0.1 wt\% C and a stoichiometric amount of Ti, which was investigated by Depover et al. \citep{depover_effect_2016}. Before TDS testing, the alloy was quenched to obtain a full martensitic structure and subsequently subjected to tempering at $600^\circ\text{C}$ for $1\;\text{h}$, for the generation of TiC precipitates. The same protocol as in the previous test case for defining the parameters is followed. The thickness of the sample and heating rate were defined as $L=0.001\;\text{m}$ and $600^{\circ}\text{C}/\text{h}$ ($\phi=0.167$ K/s), respectively, as specified in the original study. In terms of the other test parameters, the resting time was taken as $\text{t}_\text{rest}=3600\;\text{s}$. Finally, the minimum and maximum temperatures were defined as $T_\text{min}=293.15\;\text{K}$ and $T_\text{max}=873.15\;\text{K}$. Given that the material in question is a tempered martensitic steel, the relevant lattice properties were taken as the properties characteristic of the bcc lattice (\cref{table:Material_Properties}). 

The TDS spectra of tempered martensitic samples are often characterised by a low-energy high-density trap, which can be attributed to the sum of several defects with similar $\Delta H$, such as dislocations and martensitic lath boundaries \citep{drexler_model-based_2019}. The data generation process was altered to account for these types of spectra. Simply expanding the allowable trap binding energy and density ranges is not a viable option. It was observed that model performance severely degrades when dealing with very large density ranges\textemdash particularly those exceeding two orders of magnitude. Additionally, due to the random nature of the trap characteristic generation process, there is a possibility that traps with low binding energies and low densities are generated. In such cases, there is a risk that the hydrogen from these traps desorbs during the rest period. As a result, these traps will not have a corresponding peak in the TDS spectrum, which can have an adverse impact on model training and performance.

To improve the quality of the training data\textemdash specifically, how closely it resembles the experimental data\textemdash without compromising the accuracy and efficiency of the model, the data generation process is modified by separating the trapping characteristic generation phase into two steps. In the first step, the characteristics of the low-energy, high-density trap, also referred to as the first trap, are randomly generated. In the second step, the characteristics of the remaining traps are defined. This separation ensures the correct energy and density ranges are assigned for the random generation of each trap. The data generation parameters for each step were chosen by selecting a realistic range of trapping binding energies and densities that would produce fluxes consistent with the magnitudes observed in the TDS spectra. For the first trap, the absolute binding energy range is set between $30\;\text{kJ/mol}$ and $50\;\text{kJ/mol}$, with the density range between $40$ and $100\;\text{mol/m}^3$ ($2.409\times10^{225}-6.022\times10^{25}\;\text{sites/m}^{3}$). For the other traps, the absolute binding energy range is set between $50\;\text{kJ/mol}$ and $110\;\text{kJ/mol}$, with densities ranging from $0.1\;\text{mol/m}^3$ to $10\;\text{mol/m}^3$ ($6.022\times10^{22}-6.022\times10^{24}\;\text{sites/m}^{3}$). The minimum difference between binding energies was set to $10\;\text{kJ/mol}$, and the ML models were trained for a maximum number of 4 traps. The ML models were first trained with 50,000 data points; subsequently, the analysis was repeated with 100,000 data points. Doubling the number of training data points had no significant effect, as discussed below. The same numerical simulation parameters and fitting procedure required for the conventional optimisation approach were adopted as in the previous test case.

\begin{figure*}[t]
    \centering
    \begin{subfigure}{0.32\textwidth}
        \centering
        \includegraphics[width=\linewidth]{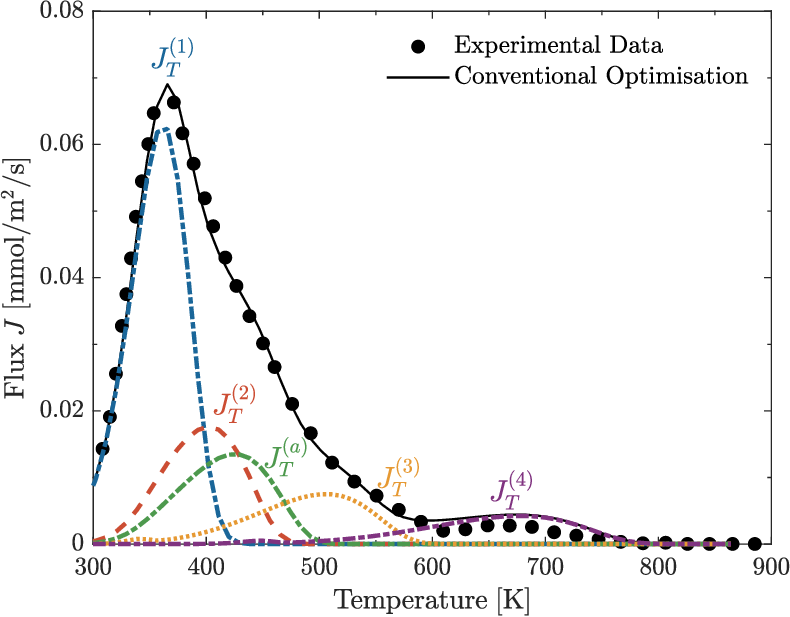}
        \caption{}
        \label{fig:Depover_Conventional}
    \end{subfigure}
    \hfill
    \begin{subfigure}{0.32\textwidth}
        \centering
        \includegraphics[width=\linewidth]{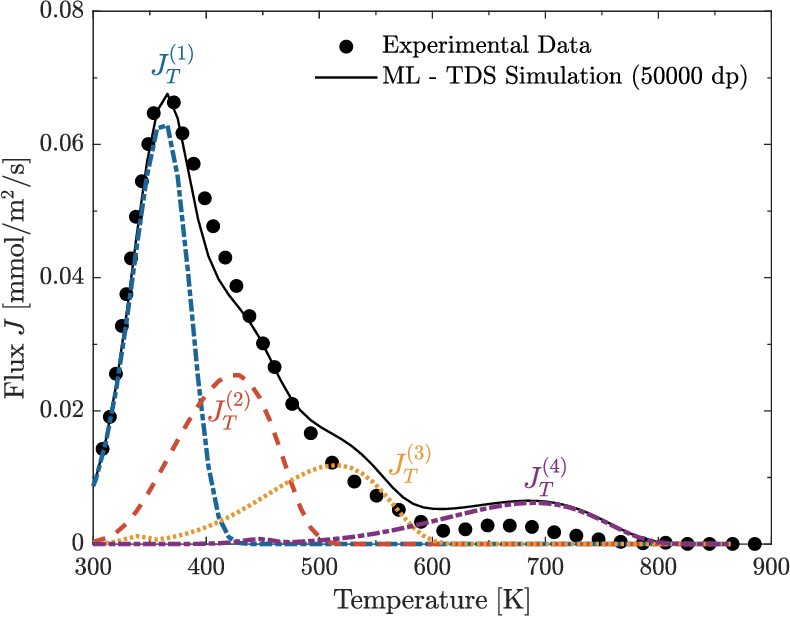}
        \caption{}
        \label{fig:Depover_ML_50}
    \end{subfigure}
    \hfill
    \begin{subfigure}{0.32\textwidth}
        \centering
        \includegraphics[width=\linewidth]{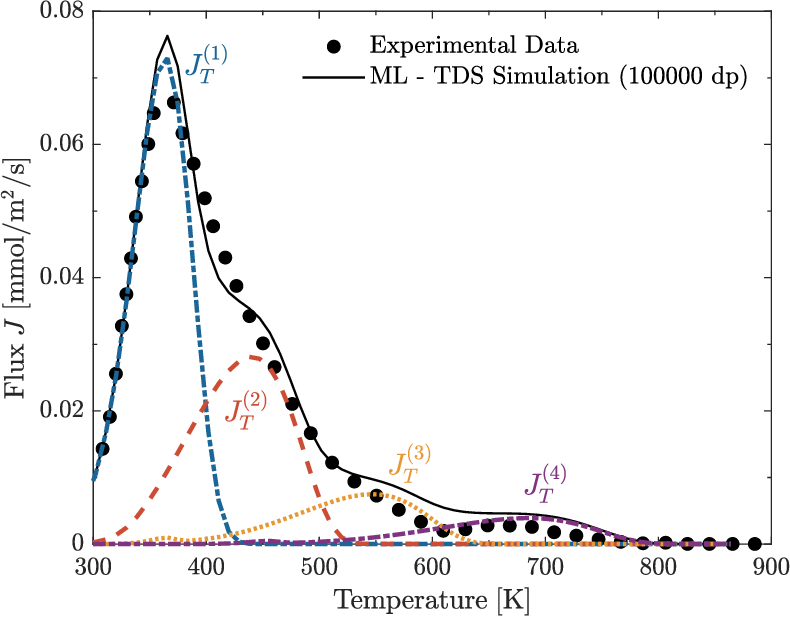}
        \caption{}
        \label{fig:Depover_ML_100}
    \end{subfigure}
    \caption{Gaining insight into the trapping characteristics of a tempered martensitic Fe-C-Ti alloy. Comparison between experimental \citep{depover_effect_2016} and simulated desorption curves, with the latter being obtained using the trapping parameters as determined by (a) conventional optimisation (\texttt{TDS Simulator} \citep{garcia-macias_tds_2024}), (b) ML models trained with 50,000 data points and (c) ML models trained with 100,000 data points. The contributions of each relevant trap type $i$, denoted $J_T^{(i)}$, as determined by each respective approach, are also illustrated. The traps are reported in order of absolute binding energy $|\Delta H|$. The additional trap identified by the conventional optimisation (\texttt{TDS Simulator}) approach is referred to as $J_T^{(a)}$.}
    \label{fig:Depover_Predictions}
\end{figure*}

The conventional optimisation and ML approach predictions are illustrated in \cref{fig:Depover_Predictions}, showing both the simulated desorption curves and individual trap contributions. Specifically, \cref{fig:Depover_Conventional} presents the reconstructed spectrum based on trapping parameters fitted via the conventional optimisation approach, while \cref{fig:Depover_ML_50} and \cref{fig:Depover_ML_100} show the spectra reconstructed from the trapping parameters predicted by ML models trained on 50,000 and 100,000 data points, respectively. The results obtained by each approach show similar qualitative trends but differ quantitatively, particularly in the number of traps identified. The trapping parameters inferred by the conventional optimisation approach reveal that the desorption curve can be deconvoluted into five peaks, each associated with a distinct trap type. Conversely, both ML analyses report only four trap types. Each trap predicted by the ML models can be matched to a trap inferred by the conventional optimisation approach, considering similarities in peak temperature and height. These traps are reported in order of absolute binding energies $|\Delta H|$ (see \cref{fig:Depover_Predictions}). The additional trap identified by the conventional optimisation approach, shown in \cref{fig:Depover_Conventional}, is referred to as Trap a. 

Reconstructing the TDS spectrum based on the fitted parameters obtained using the conventional optimisation approach produced an excellent match, capturing all major features of the desorption curve (see \cref{fig:Depover_Conventional}). The most prominent peak is governed by a low-energy ($\Delta H^{(1)}=-36.9\;\text{kJ/mol}$) high density ($N_\text{T}^{(1)}=4.33\times 10^{25}\;\text{sites/m}^3$) trap, which is attributed to a collection of defects with similar binding energies, typically associated with the martensitic microstructure (e.g., lath boundaries and dislocations) \citep{drexler_model-based_2019}. The smooth drop in the desorption rate following this peak is caused by three traps, linked to carbon vacancies at the carbide/matrix interface \citep{drexler_model-based_2019}. Two of these three traps share very similar characteristics: binding energies of $\Delta H^{(2)}=-56.2\;\text{kJ/mol}$ and $\Delta H^{(a)}=-61.1\;\text{kJ/mol}$, with trap densities of $N_\text{T}^{(2)}=1.41\times 10^{24}\;\text{sites/m}^3$ and $N_\text{T}^{(a)}=1.09\times 10^{24}\;\text{sites/m}^3$, respectively. The third trap in this group has a stronger binding energy ($\Delta H^{(3)}=-76.3\;\text{kJ/mol}$) and a lower trap density ($N_\text{T}^{(3)}=7.0\times 10^{23}\;\text{sites/m}^3$). The slight difference in $\Delta H$ between the former two traps and the latter one is the type of interface in which the carbon vacancies are found. Atomistic simulations were conducted by Di Stefano et al. \citep{di_stefano_first-principles_2016} to understand the interaction of hydrogen with TiC in iron and provide a more in-depth understanding of the difference in trap sites. Binding energies of $-44\;\text{kJ}/\text{mol}$ and $-87\;\text{kJ}/\text{mol}$ were reported for the (001)- and (110)-interface, respectively. Thus, Trap 3 can be attributed to the (110)-interface. Conversely, Trap 2 and a are most likely connected to the (001)-interface as they have a lower absolute $\Delta H$. Finally, the high-temperature peak observed around $700\;\text{K}$ corresponds to a deep trap with a strong binding energy of $\Delta H^{(4)}=-107.4\;\text{kJ/mol}$ and a low trap density ($N_\text{T}^{(4)}=5.10\times 10^{23}\;\text{sites/m}^3$). This trap can be associated with carbon vacancies within the TiC precipitates \citep{drexler_microstructural_2020}.

The TDS curve reconstructed from the predictions of the ML model trained with 50,000 data points exhibits good agreement with the experimental data at the low-temperature peak (see \cref{fig:Depover_ML_50}). However, notable discrepancies are observed at the high-temperature peak, where the model significantly overestimated the density. Additionally, mismatches are visible on the right-hand side of the low-temperature peak. The first trap, which dominates the prominent low-temperature peak, was identified by the ML model as having a binding energy of $\Delta H^{(1)} = -37.3\;\text{kJ/mol}$ and trap density of $N_\text{T}^{(1)} = 3.78 \times 10^{25}\;\text{sites/m}^3$, consistent with the conventional optimisation results. However, the traps responsible for the smooth drop in desorption differ from the conventional model. The ML model identified only two traps instead of three, with binding energies of $\Delta H^{(2)} = -58.7\;\text{kJ/mol}$ and $\Delta H^{(3)} = -75.8\;\text{kJ/mol}$, and densities of $N_\text{T}^{(2)} = 2.13 \times 10^{24}\;\text{sites/m}^3$ and $N_\text{T}^{(3)} = 1.14 \times 10^{24}\;\text{sites/m}^3$, respectively. The final trap, corresponding to the high-temperature peak, showed similar parameters to those from the conventional approach, with $\Delta H^{(4)} = -107.3\;\text{kJ/mol}$ and $N_\text{T}^{(4)} = 7.77 \times 10^{23}\;\text{sites/m}^3$. To improve the ML model predictions, the analysis was repeated with 100,000 training data points (see \cref{fig:Depover_ML_100}). Despite the increased volume of training data, no significant improvement in model performance is observed. Similar trapping parameters were found as for the 50,000 training data points case. While the ML model captured the more complex features of the TDS curve\textemdash particularly the regions near the high-temperature peak\textemdash it severely overestimated the density at the low-temperature peak. The limited improvement can be attributed to the data generation process, which is based on random distributions and may not adequately capture the underlying patterns required for effective learning. In ANN, increasing the quantity of training data does not guarantee enhanced model performance if the data is not sufficiently representative, as the model may struggle to extract meaningful features or generalise beyond the training data.

\begin{figure}
    \centering
    \includegraphics[width=0.5\textwidth]{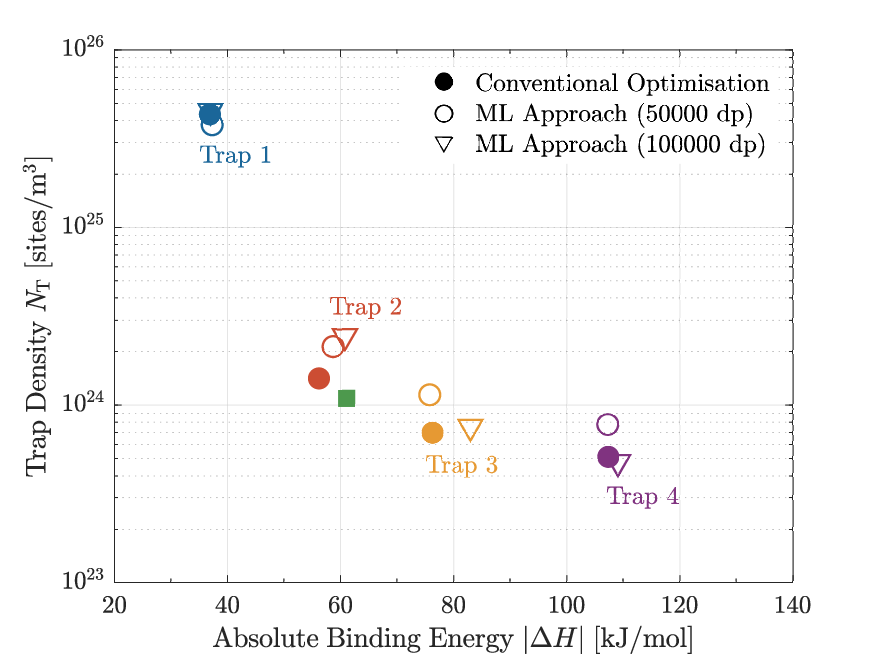}
    \caption{Scatter plot comparing the absolute binding energies ($|\Delta  H|$) and trap densities ($N_\text{T}$) obtained using conventional optimisation (\texttt{TDS Simulator} \citep{garcia-macias_tds_2024}) and ML approaches. The plot illustrates the agreement between the two approaches in predicting the trapping characteristics of a tempered martensitic Fe-C-Ti alloy from experimental data \citep{depover_effect_2016}. The traps are reported in order of absolute binding energy $|\Delta H|$. The additional trap identified by the conventional optimisation (\texttt{TDS Simulator} \citep{garcia-macias_tds_2024}) approach is depicted in green. The y-axis is shown on a logarithmic scale.}
    \label{fig:Depover_Comparison_Plot_Trapping_Parameters}
\end{figure} 

The agreement between the quantification techniques is assessed by comparing the inferred trapping parameters in \cref{fig:Depover_Comparison_Plot_Trapping_Parameters}. The traps are labelled as defined in \cref{fig:Depover_Predictions}. Strong agreement between the approaches is observed for Traps 1 and 4, while the agreement is weaker for Traps 2 and 3. The ML model's inability to adequately fit the experimental data around the peaks corresponding to Traps 2 and 3 might be partly due to the model being trained with an incorrect maximum number of traps. The conventional optimisation approach showed that the experimental TDS curve can be accurately reproduced using a total of five traps, including an intermediate trap between Traps 2 and 3 with a binding energy differing by less than $5\;\text{kJ}/\text{mol}$ from that of Trap 2 (see \cref{fig:Depover_Conventional}). In contrast, the ML model was limited to a maximum of four traps and a minimum binding energy difference of $5\;\text{kJ}/\text{mol}$, preventing it from accurately deconvoluting this region into three separate traps. Although increasing the maximum number of traps in the ML model could potentially improve the experimental data fit, doing so is currently impractical due to the associated computational costs. The architecture of both NNs scales with the number of outputs, so increasing from four to five traps (i.e., 8 to 10 outputs for the regression model and 4 to 5 outputs for the classification model) significantly expands the number of trainable parameters. This not only increases training time and memory requirements but also demands a larger dataset to avoid over-fitting and ensure model generalisation. Given that the current model already showed limitations when trained on 100,000 data points for four traps, a substantially larger dataset would be required for five traps, making the approach computationally expensive. 

\subsection{Tempered martensitic Fe-0.05C-0.20Ti-2.0Ni alloy}

The third test case involves the TDS spectrum of a tempered martensitic Fe-0.05C-0.20Ti-2.0Ni alloy, obtained by Wei and Tsuzaki \citep{wei_hydrogen_2003}. The values reported by the authors were used to define the test parameters. The thickness of the sample and heating rate were taken as $L=0.005\;\text{m}$ and $100^{\circ}\text{C}/\text{h}$ ($\phi=0.0278$ K/s), respectively. The resting time, not specified by the authors, was estimated as $\text{t}_\text{rest}=120\;\text{s}$ based on the initial drop in desorption rate observed in the spectrum. Finally, the minimum and maximum temperatures were defined as $T_\text{min}=293.15\;\text{K}$ and $T_\text{max}=1200\;\text{K}$. Similar to the previous two test cases, material parameters were taken from the properties characteristic of the ferritic lattice, see \cref{table:Material_Properties}, as the material under investigation is a tempered martensitic steel. Given that Wei and Tsuzaki \citep{wei_hydrogen_2003} use a recombination poison ($\text{NH}_4\text{SCH}$) to enhance hydrogen uptake, an initial lattice hydrogen concentrations $C_L^0$ ten times greater than that reported for ferritic lattices ($C_L^0=0.6\;\text{mol}/\text{m}^3$) is employed in this study. 

Similar to test case two, the alloy in question contains Ti carbides. However, comparing the spectrum of the Fe-0.05C-0.20Ti-2.0Ni alloy to that of the Fe-C-Ti alloy of the former test case, key differences can be observed. The most critical difference is the initial drop in desorption rate in the spectrum of the Ti carbide containing steel, which is absent in that of the Fe-C-Ti alloys. This spike is due to rapid desorption and originates from the presence of a high-density, low-energy trap. It is a direct consequence of the short resting time adopted by Wei and Tsuzaki \cite{wei_hydrogen_2003}. To account for this low-energy, high-density trap, the data generation process is modified by separating the trapping characteristic generation phase into two steps, following the procedure described in the previous case study (Section \ref{Sec:CaseStudy2}). For the first trap, the range for absolute binding energies was set between $10$ and $20\;\text{kJ/mol}$, while trap densities were constrained to a range of $1\times10^4\;-\;1\times10^5\;\text{mol}/\text{m}^3$ ($6.022\times10^{27}-6.022\times10^{28}\;\text{sites/m}^{3}$). For the remaining traps, the range for absolute binding energies was set between $40$ and $140\;\text{kJ/mol}$, with a minimum difference between binding energies of $10\;\text{kJ/mol}$. Trap densities were constrained to a range of $1-20\;\text{mol}/\text{m}^3$ ($6.022\times10^{23}-1.2044\times10^{25}\;\text{sites/m}^{3}$). The more restricted density range of these traps was introduced to minimise any interference with model performance, as it was observed in the previous case studies that density ranges spanning more than two orders of magnitude cause a substantial deterioration in the model's predictive capability. Relative to the previous case studies, a greater magnitude of Gaussian noise, with a standard deviation of 0.1, was necessary to improve the quality of the fit. The ML models were trained with $50,000$ data points to predict a maximum of $4$ traps. In regard to the conventional optimisation approach, the default numerical simulation and fitting parameters were used, as in the previous case studies.

\begin{figure*}[t]
    \centering
    \begin{subfigure}{0.45\textwidth}
        \centering
        \includegraphics[width=\textwidth]{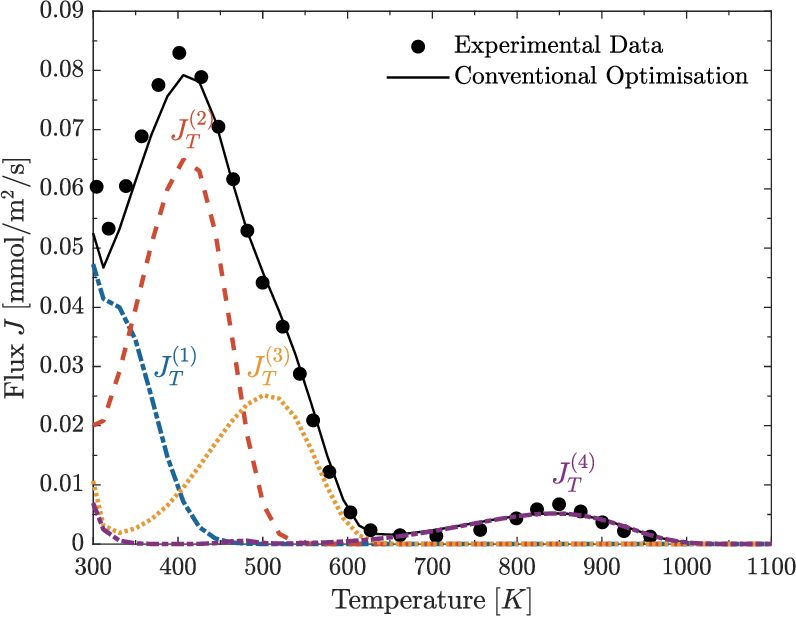}
        \caption{}
        \label{fig:Wei_Conventional}
    \end{subfigure}
    \hfill
    \begin{subfigure}{0.45\textwidth}
        \centering
        \includegraphics[width=\textwidth]{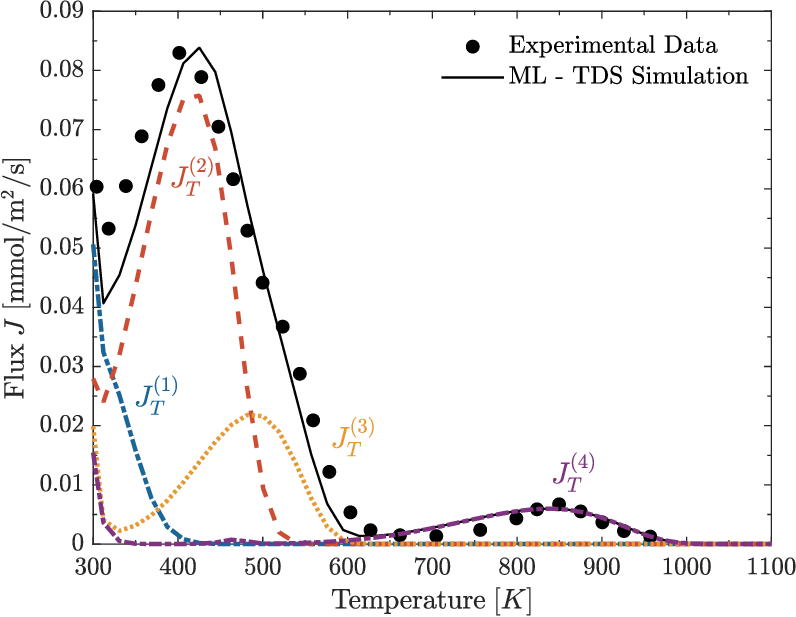}
        \caption{}
        \label{fig:Wei_ML}
    \end{subfigure}
    \caption{Gaining insight into the trapping characteristics of a tempered martensitic Fe-0.05C-0.20Ti-2.0Ni alloy. Comparison between experimental \citep{wei_hydrogen_2003} and simulated desorption curves, with the latter being obtained using the trapping parameters as determined by (a) conventional optimisation (\texttt{TDS Simulator} \citep{garcia-macias_tds_2024}), and (b) the present ML approach. The contributions of each relevant trap type $i$, denoted $J_T^{(i)}$, as determined by each respective approach, are also illustrated. The traps are reported in order of absolute binding energy $|\Delta H|$.}
    \label{fig:Wei_Predictions}
\end{figure*}

The results of the conventional optimisation and ML analyses are given in \cref{fig:Wei_Predictions}. First, \cref{fig:Wei_Conventional} shows the simulated desorption curve based on the conventional optimisation predictions alongside experimental data. As in the previous test cases, the conventional optimisation approach is able to produce an excellent fit, capturing all major features of the TDS curve. The analysis reveals that the spectrum can be deconvoluted into four peaks, each corresponding to a distinct trap. Similar results are obtained by the ML approach as shown in \cref{fig:Wei_ML}. The ML approach is able to capture the characteristics of the desorption spectra very well, from the initial drop in desorption rate to the smallest ones that appear at higher temperatures. A slight mismatch is observed between the experimental data and ML model predictions at the initial drop in desorption rate. As in the conventional optimisation approach, the response is captured by four traps. Good agreement is observed between the two approaches. Comparing the trap contributions in each spectrum (see \cref{fig:Wei_Predictions}), similar qualitative trends can be observed. Specifically, the peak temperature corresponding to each trap type is almost identical across both approaches. Additionally, the order of trap type densities is the same for all spectra ($\text{Trap 1}>\text{Trap 2}>\text{Trap 3}>\text{Trap 4}$). 

\begin{figure}
    \centering
    \includegraphics[width=0.5\textwidth]{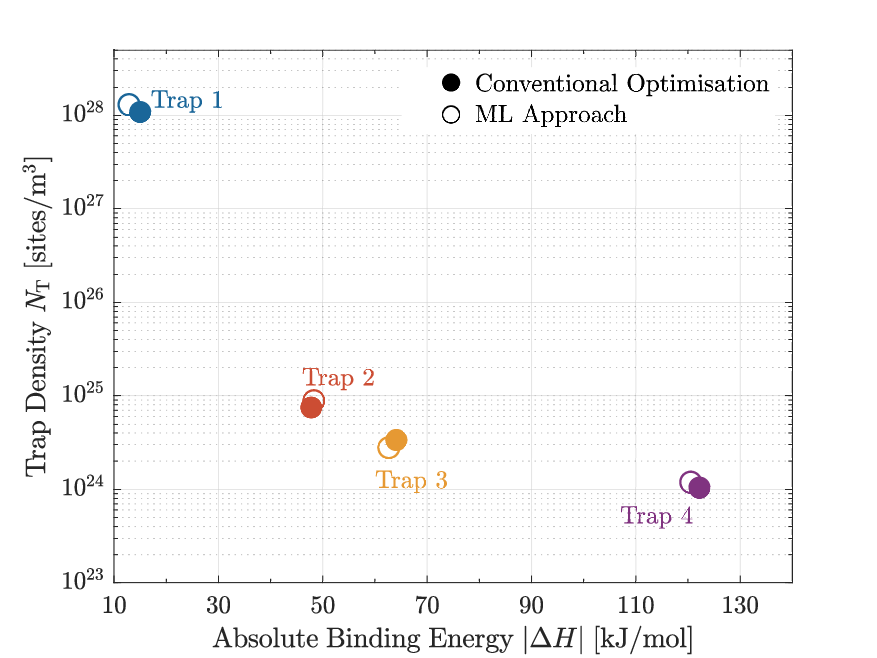}
    \caption{Scatter plot comparing the absolute binding energies ($|\Delta  H|$) and trap densities ($N_\text{T}$) obtained using conventional optimisation (\texttt{TDS Simulator} \citep{garcia-macias_tds_2024}) and ML approaches. The plot illustrates the agreement between the two approaches in predicting the trapping characteristics of a tempered martensitic Fe-0.05C-0.20Ti-2.0Ni alloy from experimental data \citep{wei_hydrogen_2003}. The traps are reported in order of absolute binding energy $|\Delta H|$. The y-axis is shown on a logarithmic scale. }
    \label{fig:Wei_Comparison_Plot_Trapping_Parameters}
\end{figure}

Subsequent analysis and comparison of the trapping parameters inferred by each approach provide greater insight into the trapping characteristics of the tempered martensitic Fe-0.05C-0.20Ti-2.0Ni alloy, as well as the agreement between the two quantification techniques (shown in \cref{fig:Wei_Comparison_Plot_Trapping_Parameters}). Very similar trap densities and binding energies are reported by the approaches, specifically for Traps 2 to 4. A larger discrepancy is observed in both trapping parameters of Trap 1. Both approaches attribute the initial spike in desorption to a shallow trap (Trap 1) with a significantly higher density, by four orders of magnitude, compared to the other traps. The conventional optimisation method assigns this trap a binding energy of $\Delta H^{(1)}=-15.0\;\text{kJ/mol}$ and a density of $N_\text{T}^{(1)}=1.08\times10^{28}\;\text{sites/m}^3$, while the ML approach predicts a slightly lower binding energy of $\Delta H^{(1)}=-12.9\;\text{kJ/mol}$ and a higher density of $N_\text{T}^{(1)}=1.30\times10^{28}\;\text{sites/m}^3$. The difference of $3\;\text{kJ/mol}$ in binding energy between the two methods helps explain the mismatch between the ML-predicted and experimental curves observed in \cref{fig:Wei_ML} at the initial desorption drop. The prominent low-temperature peak in the desorption data was found to result from the presence of two traps (Traps 2 and 3). Using the conventional optimisation approach, their binding energies and densities were determined as $\Delta H^{(2)}=-47.8\;\text{kJ/mol}$ and $\Delta H^{(3)}=-64.1\;\text{kJ/mol}$, and $N_\text{T}^{(2)}=7.50\times10^{24}\;\text{sites/m}^3$ and $N_\text{T}^{(3)}=3.36\times10^{24}\;\text{sites/m}^3$, respectively. In comparison, the ML approach predicted $\Delta H^{(2)}=-48.3\;\text{kJ/mol}$ and $\Delta H^{(3)}=-62.7\;\text{kJ/mol}$, and $N_\text{T}^{(2)}=8.85\times10^{24}\;\text{sites/m}^3$ and $N_\text{T}^{(3)}=2.78\times10^{24}\;\text{sites/m}^3$ for the same traps. Based on typical values reported in the literature, these traps can be attributed to grain boundaries, dislocations, and coherent $\text{TiC}$. Finally, the high-temperature peak located around $800-900\;\text{K}$ is attributed to a deep trap (Trap 4) associated with incoherent $\text{TiC}$ particles. Similar binding energies of $\Delta H^{(4)}\;=-122.2\;\text{kJ/mol}$ and $\Delta H^{(4)}=-120.5\;\text{kJ/mol}$ and trap densities of $N_\text{T}^{(4)}=1.04\times10^{24}\;\text{sites/m}^3$ and $N_\text{T}^{(4)}=1.19\times10^{24}\;\text{sites/m}^3$ were reported by the conventional optimisation and ML approach, respectively. 

\subsection{Comparison of the trapping characteristics of tempered martensitic steels}

A comparison of the trapping characteristics predicted by the newly developed ML approach for each tempered martensitic steel is presented in \cref{table:comparison_traps_test_cases}. Due to the microstructural similarities among the three alloys, several common trapping features were observed. Notably, all steels exhibited a low-energy, high-density trap\textemdash identified as Trap 1 for the high-strength AISI 4340 and Fe–C–Ti alloys (Test Cases 1 and 2), and as Traps 1 and 2 for the Fe–0.05C–0.20Ti–2.0Ni alloy (Test Case 3). These traps are likely associated with typical martensitic features such as dislocations and lath boundaries \citep{wei_hydrogen_2003, drexler_model-based_2019, chen_hydrogen_2025}. Additionally, all alloys showed a trap in the $60–75$ kJ/mol binding energy range, most likely corresponding to coherent carbides or carbide–matrix interfaces \citep{di_stefano_first-principles_2016, drexler_model-based_2019, chen_hydrogen_2025}. A key distinction was observed in Test Case 3, which exhibited a significantly stronger trapping site with a binding energy of $120.5$ kJ/mol \citep{wei_hydrogen_2003}. This high-energy trap is attributed to the presence of incoherent TiC particles, which are not found in the other two alloys.

\begin{table*}[h!]
    \centering
    \setlength{\tabcolsep}{3pt} % Horizontal padding
    \renewcommand{\arraystretch}{1} % Vertical padding
    \begin{tabular}{lccccc}
        \toprule
        \textbf{} & & \textbf{Trap 1} & \textbf{Trap 2} & \textbf{Trap 3} & \textbf{Trap 4} \\
        \midrule

        \multirow{2}{*}{\textbf{Test Case 1}} 
            & $|\Delta H|$ (kJ/mol) 
            & $48.7$--$52.6$ & $61.5$--$74.7$ & $90.3$--$95.7$ & $128.9$--$144.8$ \\
        & $N_\text{T}$ (sites/m$^3$) 
            & $(5.4$--$6.1)\times 10^{24}$ & $(0.70$--$1.5)\times 10^{24}$ 
            & $(4.6$--$9.1)\times 10^{23}$ & $(0.67$--$1.3)\times 10^{24}$ \\
        \midrule

        \multirow{2}{*}{\textbf{Test Case 2}} 
            & $|\Delta H|$ (kJ/mol) 
            & $37.0$--$37.3$ & $58.7$--$60.8$ & $75.8$--$83.0$ & $107.3$--$109.1$ \\
        & $N_\text{T}$ (sites/m$^3$) 
            & $(3.8$--$4.5)\times 10^{25}$ & $(2.1$--$2.4)\times 10^{24}$ 
            & $(0.75$--$1.1)\times 10^{24}$ & $(4.8$--$7.8)\times 10^{23}$ \\
        \midrule

        \multirow{2}{*}{\textbf{Test Case 3}} 
            & $|\Delta H|$ (kJ/mol) 
            & $12.9$ & $48.3$ & $62.7$ & $120.5$ \\
        & $N_\text{T}$ (sites/m$^3$) 
            & $1.3\times 10^{28}$ & $8.9\times 10^{24}$ 
            & $2.8\times 10^{24}$ & $1.2\times 10^{24}$ \\
        \bottomrule
    \end{tabular}
    \caption{Comparison of the hydrogen trapping characteristics, i.e., the absolute binding energies and trap densities, extracted using the novel machine learning approach for three tempered martensitic steels with varying compositions. Test Case 1 corresponds to high-strength AISI 4340 steel \citep{novak_statistical_2010}, Test Case 2 to a Fe-C-Ti alloy \citep{depover_effect_2016}, and Test Case 3 to a Fe–0.05C–0.20Ti–2.0Ni alloy \citep{wei_hydrogen_2003}}
    \label{table:comparison_traps_test_cases}
\end{table*}

\subsection{Limitations}
\label{sec:limitations}

While the ML models provide good fits to the experimental datasets, the conventional optimisation approach results in more accurate predictions across all test cases. This is primarily due to the ability of the PSO algorithm to refine its predictions with increasing runtime. In contrast, once trained, the ML models provide a single prediction for a given input. Improving the predictions requires retraining the model with an increased volume of training data. Even then, retraining does not always guarantee improved performance. Moreover, the dataset and training used in the ML approach are specific to the TDS test and material parameters (e.g., lattice diffusivity, heating rate, sample thickness). As a result, each ML model is not generalizable to experimental datasets that differ significantly from the conditions it was trained on. However, TDS parameters (e.g., sample size) typically remain consistent across experiments, and the relevant material parameters (e.g., lattice diffusivity) are common to a wide range of materials (being lattice properties). As such, the training of the present TDS-ML approach on representative test parameters and material classes (fcc, bcc) would enable efficient predictions across alloys within the same class - e.g., quantifying the role of micro-structural and compositional changes on trapping characteristics. Additionally, since training and data generation can be completed before the experiments, the model allows for rapid, near-instantaneous processing of experimental data. For certain applications, the ML approach offers clear advantages over the conventional optimisation approach, which is time-consuming, as optimisation must be performed separately for each experiment, and may not reach the optimal solution if the optimisation algorithm encounters a local minimum.

However, there are other limitations associated with the ML approach that must also be taken into consideration. One additional challenge encountered was the sensitivity of the ML model to features poorly captured by the TDS spectrum. If the sample rest time (the time the sample is kept at room temperature before the start of the temperature ramp) is too short, part of the spectrum captures the diffusion of lattice hydrogen out of the sample, creating a peak at the start of the spectrum. As this peak is not uniquely defined (i.e., many possible combinations of trapping energy/density can result in a similar peak), the ML approach has issues properly learning how to identify the trapping energies and densities. 

One final limitation of the ML approach is related to the range of trapping densities used within the model. Despite the scaling employed in the ML model, the scheme was only able to predict trapping densities across two orders of magnitude. This is also due to the limitations of pre-trained artificial neural networks. If extremely high and low densities are combined within a single model, the low-density peaks are overshadowed by the high-density peaks. As a result, the model is able to learn to identify the high-density trapping parameters, but is unable to identify the low-density traps. 

Finally, it is essential to acknowledge the limitations inherent in the constitutive model used to generate the training data. The model is based on the McNabb-Foster and Oriani formulations, both of which assume isolated, sparsely distributed traps and therefore neglect any potential interactions between traps. While this assumption is generally valid for BCC materials due to their high diffusivity \citep{simoni_integrated_2021, drexler_model-based_2019, song_theory_2013}, it becomes less appropriate for materials with lower diffusivity, such as FCC alloys, where trapping behaviour is influenced by the distribution, morphology and interconnectivity of the traps \citep{wang_influence_2016}. In such cases, these interactions cannot be disregarded. Extending the present approach to these materials will require incorporating more complex trap interactions in the numerical model as well as accounting for microstructural effects.

\section{Conclusions}
We have presented a neural network-based machine-learning model able to identify the trapping characteristics from thermal desorption experimental data. The model has been validated against the existing literature and its potential has been assessed through multiple case studies, extensively discussing its limitations and capabilities. The main advantages of this approach, relative to state-of-the-art TDS optimisation fitting approaches, are:
\begin{itemize}
    \item The model is trained on purely numerical simulations, allowing it to be generated without needing any experimental inputs. This allows the models to be trained for a wide range of circumstances, providing an easy and quick way to process TDS curves to extract the trapping characteristics. 
    \item The multi-ANN approach allows the number of unique traps to be first determined, and then automatically predicts the appropriate energies and densities for these trapping sites.
    \item In contrast to conventional optimisation schemes, the machine-learning-based model does not have any issues with convergence related to local minima, and is able to re-use the generated simulation data between studies, making it a computationally cheaper method to quantify a series of TDS curves.
\end{itemize}
Among others, the results obtained show that,
\begin{itemize}
    \item The model can efficiently find trapping characteristics (number of trap types, along with their densities and binding energies) that closely match experimental TDS spectra and that are in good agreement with the fitting resulting from conventional optimisation approaches
    \item Addition of Gaussian noise to the training data is essential for accurately fitting experimental data, as synthetic data is noise-free, while real-world data is inherently noisy.  
    \item Despite optimised input scaling techniques being employed in the model, the limitations of pre-trained neural networks and the ill-posed nature of the TDS analysis restrict the framework to predicting trap densities over a range of two orders of magnitude.
\end{itemize}
Future work will be aimed at overcoming the limitations of the model, discussed in \cref{sec:limitations}, so as to further enhance its predictive capabilities. A primary goal will be to improve the scalability of the approach, enabling the development of a model applicable to an entire material class (e.g., bcc alloys) and typical TDS test parameters. 

\section*{Data availability}
\noindent The finite element simulation code and machine learning approaches employed are available at \url{https://github.com/nicolettamarrani/TDS_ML_Approach.git}. Example inputs and files for post-processing, allowing the results from Figs. 8-9 to be reproduced, are also included.

\section*{Declaration of competing interest}
\noindent The authors declare that they have no known competing financial interests or personal relationships that could have appeared to influence the work reported in this paper.

\section*{Acknowledgements}
\noindent N. Marrani and E. Martinez-Pañeda acknowledge support through the UKRI Horizon Europe Guarantee programme (ERC Starting Grant \textit{ResistHfracture}, EP/Y037219/1). T. Hageman acknowledges support via the Royal Commission for the Exhibition of 1851 fellowship scheme and the John Fell Oxford University Press Research Fund. E. Martinez-Pañeda also acknowledges support from UKRI's Future Leaders Fellowship programme [grant MR/V024124/1]. The authors also acknowledge computational resources and support provided by the University of Oxford Advanced Research Computing Service (\url{http://dx.doi.org/10.5281/zenodo.22558}).

\section*{CRediT authorship contribution statement}
\textbf{N. Marrani}: Methodology, Software, Validation, Formal analysis, Investigation, Visualisation, Data Curation, Writing - Original Draft \textbf{T. Hageman}: Methodology, Resources, Writing - Original Draft - Review \& Editing, Supervision \textbf{E. Martínez-Pañeda}: Conceptualisation, Resources, Writing - Review \& Editing, Supervision, Project administration, Funding acquisition

\FloatBarrier

\end{document}